\documentclass{article}

\usepackage[preprint]{neurips_2025}

\usepackage[utf8]{inputenc} 
\usepackage[T1]{fontenc}    
\usepackage{hyperref}       
\usepackage{url}            
\usepackage{booktabs}       
\usepackage{amsfonts}       
\usepackage{nicefrac}       

\usepackage {mathtools,amssymb,amsthm} 
\usepackage{csquotes}
\usepackage{cleveref}
\usepackage{enumerate}
\usepackage{graphicx}
\usepackage{multirow}
\usepackage{tabularx}
\usepackage{array}
\usepackage{enumitem}
\usepackage[ruled,vlined]{algorithm2e}

\usepackage{booktabs}
\usepackage{tcolorbox}
\tcbuselibrary{most}
\usepackage{listings}
\usepackage[edges]{forest}
\usetikzlibrary{shadows}
\usepackage{caption}
\usepackage{lipsum} 
\usepackage[export]{adjustbox} 
\usepackage{setspace}        
\tcbuselibrary{skins,breakable} 

\usepackage{algpseudocode}
\usepackage{float}

\title{PIPE: Physics-Informed Position Encoding for Alignment of Satellite Images and Time Series}

\author{%
  $\textbf{Haobo Li}^1 \;\; \textbf{Eunseo Jung}^1 \;\; \textbf{Zixin Chen}^1 \;\; \textbf{Zhaowei Wang}^1$ \\ $\textbf{Yueya Wang}^2 \;\; \textbf{Huamin Qu}^1 \;\; \textbf{Alexis Kai Hon Lau}^2$ \\
  $^1$ Department of Computer Science \& Engineering \\
  $^2$ Division of Environment \& Sustainability \\
  Hong Kong University of Science and Technology\\
  \texttt{hliem@connect.ust.hk} \\
}

\begin{document}

\maketitle

\begin{abstract}
Multimodal time series forecasting is foundational in various fields, such as utilizing satellite imagery and numerical data for predicting typhoons in climate science. 
However, existing multimodal approaches primarily focus on utilizing text data to help time series forecasting, leaving the visual data in existing time series datasets untouched. 
Furthermore, it is challenging for models to effectively capture the physical information embedded in visual data, such as satellite imagery's temporal and geospatial context, which extends beyond images themselves.
To address this gap, we propose \textbf{p}hysics-\textbf{i}nformed \textbf{p}ositional \textbf{e}ncoding (\textbf{PIPE}), a lightweight method that embeds physical information into vision language models (VLMs). \textbf{PIPE} introduces two key innovations: (1) a physics-informed positional indexing scheme for mapping physics to positional IDs, and (2) a variant-frequency positional encoding mechanism for encoding frequency information of physical variables and sequential order of tokens within the embedding space. By preserving both the physical information and sequential order information, \textbf{PIPE} significantly improves multimodal alignment and forecasting accuracy. Through the experiments on the most representative and the largest open-sourced satellite image dataset, \textbf{PIPE} achieves state-of-the-art performance in both deep learning forecasting and climate domain methods, demonstrating superiority across benchmarks, including a 12\% improvement in typhoon intensity forecasting over prior works.
Our code is provided in the supplementary material.
\end{abstract}

\section{Introduction}

Time series forecasting plays a crucial role in climate modeling~\cite{yuan2020self}.
This task involves modeling temporal dependencies to predict future values of a target variable, a challenge exacerbated by noise, non-stationarity, and the frequent need to integrate heterogeneous auxiliary data. While traditional methods like Autoregressive Integrated Moving Average (ARIMA) rely on statistical priors~\cite{ho1998use}, deep learning architectures (e.g., LSTMs~\cite{hochreiter1997long}, Transformers~\cite{vaswani2017attention}) have recently dominated the field by learning latent temporal patterns from data. However, these methods still struggle to deliver precise forecasts amid the complexity and scale of real-world data, leaving high-stakes tasks such as typhoon-track prediction continue to have a long way to go.


The rise of large language models (LLMs) as a type of sequence modeling has introduced new opportunities for time series forecasting. Although LLMs were originally built for NLP tasks such as text generation~\cite{openai2024gpt4technicalreport} and summarization~\cite{dagdelen2024structured}, their core objective naturally aligns with time-series forecasting: predicting the next token in a sentence mirrors forecasting the next value in a sequence, both conditioned on historical context. 
Consequently, existing work adapts LLMs to forecasting through tokenization techniques or patching technology to splice time-series segments into model context~\cite{zhou2023one}. More recent work broadens the paradigm by injecting auxiliary instructions or descriptions through zero-/few-shot inference~\cite{chang2025llm4ts}, in-context learning~\cite{liu2024autotimes}, and text-augmented forecasting~\cite{jin2023time}. Several studies push the scope further by incorporating explicit temporal cues, for example, TimeLLM~\cite{jin2023time} and UniTime~\cite{liu2024unitime} involve temporal information in prefix-prompts, while AutoTimes embeds timestamps as positional encodings to integrate the temporal information~\cite{liu2024autotimes}.

\begin{figure}[t]
  \centering
  \includegraphics[width=\linewidth]{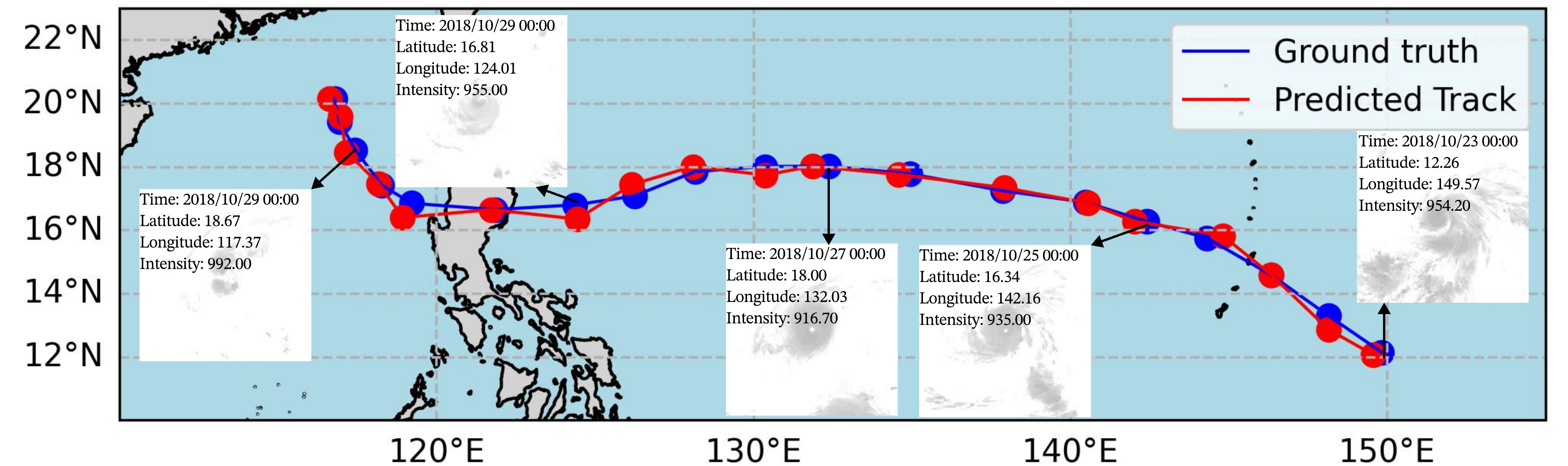}
  \caption{The multimodal time series forecasting task and the forecasting results for Typhoon Yutu by our PIPE-3B. The leading time is 12 hours and the time gap between neighbouring dots is 12 hours. In multimodal time series forecasting, satellite images can improve the forecasting accuracy.}
  \label{fig:track}
\end{figure}

However, existing methods for multimodal time series forecasting, which integrate visual and numerical data, face numerous limitations.
Integrating visual context, such as satellite imagery, into forecasting is indispensable in climate~\cite{veillette2020sevir} and other domains~\cite{jean2016combining, wang2023contrast}, yet state-of-the-art vision–language models (VLMs) like GPT-4o \cite{openai2024gpt4technicalreport}, Gemini \cite{team2023gemini}, and Qwen-VL \cite{Qwen2.5-VL} are tuned primarily for general domain multimodal data. Furthermore, their projection layers, the vision encoder from CLIP~\cite{radford2021learning}, cross-attention~\cite{lin2022cat}, Q-Former~\cite{li2023blip}, and MLP~\cite{liu2023visual}, solely focus on pixel-level semantics and overlook the rich physical metadata (e.g., timestamps and geo-coordinates) embedded in real-world imagery. This omission limits their capacity to improve high-stakes multimodal forecasting tasks. For example, typhoon track prediction with satellite imagery (\autoref{fig:track}) requires correlating pixel values with the time-specific geophysical attributes (e.g., latitude, longitude) embedded in each pixel. As a result, addressing these overlooked physical dimensions beyond the pixel-level values in multimodal time-series forecasting not only fills a critical gap in existing alignment methods that only focus on the pixel-level values but also introduces a new task for multimodal alignment.


To address these challenges, we propose \textbf{p}hysics-\textbf{i}nformed \textbf{p}ositional \textbf{e}ncoding (\textbf{PIPE}), a lightweight method to embed latent physical metadata (e.g., timestamps, geospatial coordinates) into positional encodings. Unlike traditional positional encodings, which focus solely on the sequence order of tokens within one input instance~\cite{vaswani2017attention, dosovitskiy2020image}, \textbf{PIPE} encodes shared global physical knowledge (e.g., latitude-longitude relationships consistent across instances) while preserving sequence order information. 
Specifically, \textbf{PIPE} introduces two key innovations: (1) a physics-informed positional indexing scheme that maps physics to positional IDs, and (2) variant-frequency positional encoding that integrates the attributes of physical variables in the input embedding space.
They maintain the original token topology while enabling explicit modeling of geographic-temporal dependencies.
Experiments on the most representative and the largest open-sourced satellite image dataset for typhoons, Digital Typhoon~\cite{kitamoto2023digital}, demonstrate improved cross-modal integration and forecasting accuracy. \textbf{PIPE} achieves state-of-the-art performance compared to general AI and domain models across multiple benchmarks on the multimodal time series forecasting task. 

Our contributions are threefold:

\begin{itemize}
    \item We propose the multimodal time-series forecasting framework to integrate visual information, where time series data is accompanied by corresponding vision data, extending beyond conventional univariate/multivariate time series forecasting.
    \item We propose the \textbf{PIPE}, a method to embed physical knowledge into VLMs. Our method contains two key innovations: (1) physics-informed positional indexing and (2) variant-frequency positional encoding. 
    \item Through comprehensive experiments on the most representative task and the largest open-sourced satellite image dataset, we show an obvious gain (12\% for intensity forecasting) after appropriately integrating vision and physics.
    Through the ablation study, we quantify the benefits of (1) integrating visual data for multimodal time series forecasting (8\% for intensity forecasting) and (2) integrating physics knowledge (6\% for intensity forecasting). 
\end{itemize}

\section{Related Work}

\subsection{Transformers for Time-series Forecasting}
Transformers are widely used for time series forecasting, demonstrating superior performance over traditional statistical models and RNN~\cite{rumelhart1986learning} architectures. 
Key innovations driving this success include efficient attention mechanisms and architectural adaptations tailored to temporal patterns.
Recent works have introduced several enhancements to address computational complexity and domain-specific challenges. 
Informer~\cite{zhou2021informer} tackles the quadratic complexity of standard self-attention through ProbSparse attention combined with distillation operations to prioritize crucial temporal features. 
Autoformer~\cite{wu2021autoformer} integrates decomposition from time-series analysis with autocorrelation-inspired attention and outperforms self-attention in both efficiency and accuracy.
iTransformer~\cite{liu2023itransformer} applies the attention and network on the inverted dimensions for time series forecasting.
One Fits All~\cite{zhou2023one} fine-tunes on all major types of tasks involving time series.
Other variations on transformer include CrossFormer~\cite{wang2023crossformer}, TimeXer~\cite{wang2024timexer}, TimeMixer~\cite{wang2024timemixer}, etc.

The patching paradigm has inspired multiple variants. PatchTST~\cite{nie2022time} segments time series into local windows as input tokens while maintaining channel independence for multivariate data. 
Building on this concept, works such as Pathformer~\cite{chen2024pathformer} and Sageformer~\cite{zhang2024sageformer} research transformer-based patching technology in terms of multiscale and inter-series dependencies.
Notably, works such as One Fits All~\cite{zhou2023one} and Time-LLM~\cite{jin2023time} demonstrate the transferability of patching strategies by adapting pre-trained large language models to time-series forecasting through input token alignment.

However, these developments underscore the challenge of managing complexity when incorporating additional modules, such as patch-based components. Our method incorporates physical information via Position IDs, avoiding the need for extra models.


\subsection{Multimodal LLMs for Time Series Forecasting}

Recent advances in LLMs have catalyzed efforts to develop multimodal models capable of processing diverse data modalities (e.g., text, images, audio) through unified architectures. 
This paradigm has inspired time-series forecasting adaptations that integrate textual instructions with temporal data. TimeLLM~\cite{jin2023time} reprograms the input time series with text timestamps as prefix-prompts to align the two modalities.
Unitime~\cite{liu2024unitime} utilizes prefix-prompts to encode frequency information of temporal data to augment the model.
AutoTimes~\cite{liu2024autotimes} uses the embedding of textual timestamps as the position encoding to incorporate temporal information.
Subsequent works like UrbanGPT~\cite{li2024urbangpt}, TEST~\cite{sun2024test}, ChatTime~\cite{wang2025chattime}, and GPR4MTS~\cite{jia2024gpt4mts} utilize similar methods, aligning text instructions and time series for the augmentation of time series forecasting. 

However, for time series forecasting, existing multimodal approaches focus narrowly on aligning textual instructions with numerical time series, neglecting critical vision modalities inherent to many forecasting scenarios, such as typhoon forecasting. 
Our work researches the utilization of vision data for time series forecasting.


\subsection{Position Encoding in Transformers}
Transformers require explicit position encoding to capture sequential order information, unlike RNNs that inherently model temporal relationships through hidden state propagation. Current position encoding strategies can be categorized into two primary paradigms:

1. Absolute position encodes the absolute position of a unit within a sentence.
The original Transformer architecture~\cite{vaswani2017attention} introduced two variants: 1) Learned positional embeddings during training stages.
2) Fixed sinusoidal functions: 
\begin{align}
\label{eq:sin}
    PE_{(pos,2i)}=sin(\frac{pos}{10000^{2i/d_{model}}})\\
    PE_{(pos,2i+1)}=cos(\frac{pos}{10000^{2i/d_{model}}}) \nonumber
\end{align}
where $i$ denotes the dimension, $pos$ is the position, and $d_{model}$ is the dimension of embeddings. This matrix is simply added to the embeddings before they are fed to the Transformer model.
Subsequent methods have been proposed to address the challenges of long sequences ~\cite{kitaev2020reformer, liu2020improving} and improve efficiency~\cite{press2021shortformer}.

2. Relative position encodes the position of a unit relative to other units.
Shaw et al.~\cite{shaw2018self} pioneered this approach by modifying self-attention to compute relative position biases.
Transformer-XL~\cite{dai2019transformer} introduces recurrence-aware position encoding for long-context modeling.
Ke et al.~\cite{ke2021rethinking} propose untied position embeddings to add relative position embeddings through additive scalar biases.
Wu et al.~\cite{wu2021da} propose to incorporate the real distances between tokens to re-scale the raw self-attention weights.
Rotary Position Embedding (RoPE)~\cite{su2024roformer} injects relative positions via rotation matrices.

Though effective for local sequence modeling, these methods focus on intra-instance positional relationships within individual input samples. For time-series forecasting tasks where cross-instance physical dependencies are critical (e.g., all instances share the global knowledge of geographic information), existing approaches fail to capture global temporal-spatial correlations across the entire dataset.
Our work addresses this limitation through physics-informed position encoding. By encoding global timestamps with geographic coordinates (latitude/longitude), our method preserves continuous spatiotemporal relationships across independent time-series sequences.

\section{Method}
This section formalizes the multimodal time series forecasting problem and proposes physics-informed positional encoding\textbf{PIPE}
(PIPE) that integrates physical information into
VLMs for the multimodal time series forecasting. A schematic overview of the method is provided in \autoref{fig:method}.

\subsection{Multimodal Time Series Forecasting Problem Formulation}
We address the problem of multimodal time series forecasting, where historical observations comprise both time series data of multiple variables and visual images. 
Given a sequence of historical time steps:
\begin{equation}
    \boldsymbol{x}_{t-H+1:t} = \{\boldsymbol{x}_{t-H+1}, \boldsymbol{x}_{t-H+2}, ..., \boldsymbol{x}_{t}\} \in \mathbb{R}^{H\times C}
\end{equation}
where $H$ denotes the historical time steps, $C$ the number of variates, along with a corresponding sequence of $H$ images: $\boldsymbol{i}_{t-H+1:t} \in \mathbb{R}^{3\times H_{img}\times W_{img}}$ for each time step with $H_{img}, W_{img}$ as the height and width of the image, the objective is to forecast the future $F$ time steps:
\begin{equation}
    \boldsymbol{x}_{t+1:t+F} = \{\boldsymbol{x}_{t+1}, \boldsymbol{x}_{t+2}, ..., \boldsymbol{x}_{t+F}\} \in \mathbb{R}^{F\times C}
\end{equation}
Our task is to propose a VLM model as a cross-modal forecaster $f_{VLM}(\cdot)$ to model cross-modal relationships between the multivariate sequence $\boldsymbol{x}_{t-H+1:t}$ and visual sequence $\boldsymbol{i}_{t-H+1:t}$. Formally, we seek to learn:

\begin{equation}
    \boldsymbol{\hat{x}}_{t+1:t+F} = f_{VLM}(\boldsymbol{x}_{t-H+1:t},\; \boldsymbol{i}_{t-H+1:t}) 
\end{equation}

\subsection{VLMs for Multimodal Time Series Forecasting}
To perform the multimodal time series forecasting, we use the VLM to encode the time series input and the vision input, following the practice of VLM's pipeline.

\begin{figure}[t]
  \centering
  \includegraphics[width=\linewidth]{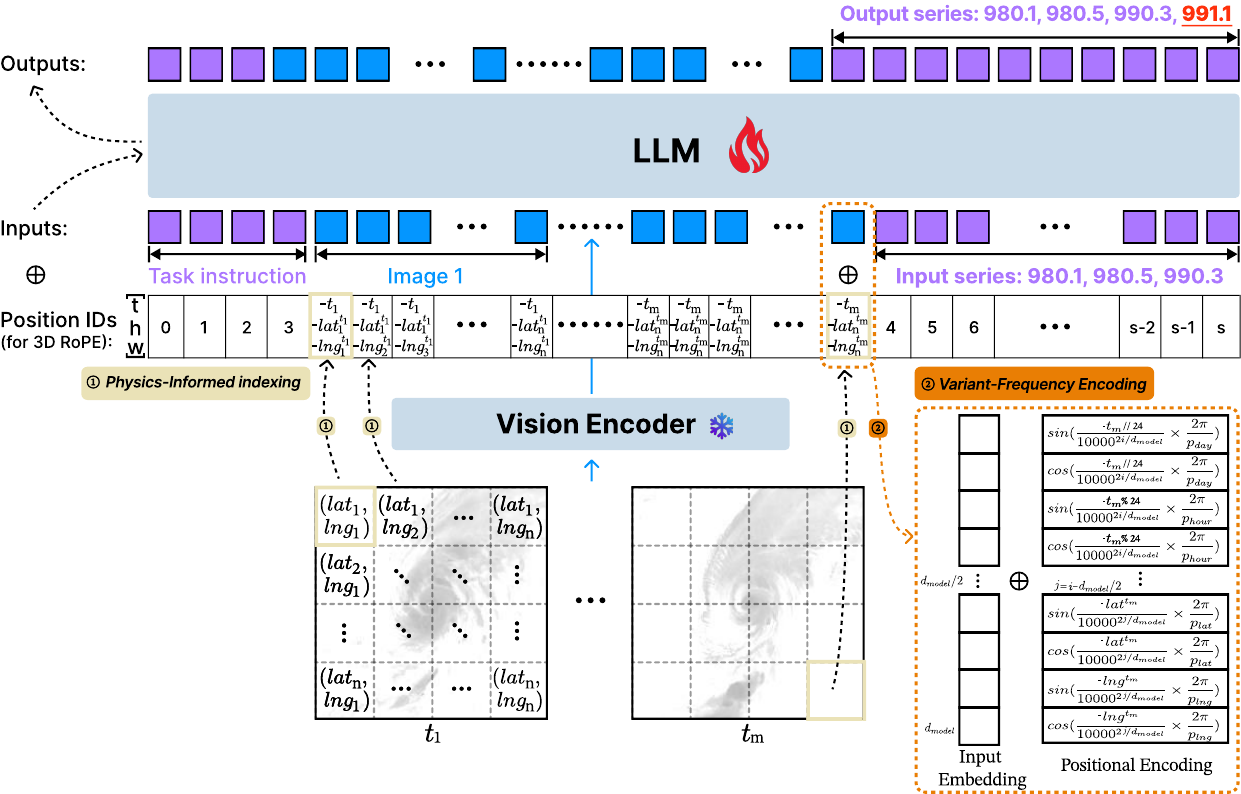}
  \caption{The framework of physics-informed positional encoding. It includes: (1) a physics-informed positional indexing scheme that maps physics to positional IDs, and (2) variant-frequency positional encoding that integrates the attributes of physical variables in the input embedding space.}
  \label{fig:method}
\end{figure}

\paragraph{Text embedding}
To leverage the capability of the pretrained LLM (Qwen-2.5-vl~\cite{Qwen2.5-VL} in this paper), we tokenize the time series data, $\boldsymbol{x}_{t-H+1:t}$, into tokens and concatenate them with task-specific instructions (e.g., ``Predict next 24 hours of typhoon track'').
They are fed into the LLM’s transformer layers, as depicted with purple inputs in \autoref{fig:method}.
The LLM is trained during the training stage.
The prompt design for multimodal time series forecasting can be found in \autoref{appendix:prompt}.

\paragraph{Vision embedding}
Each image $\boldsymbol{i}_{t} \in \mathbb{R}^{3\times H_{img}\times W_{img}}$ is split into $N$ non-overlapping patches $\{\boldsymbol{p_{t, k}}\}^N_{k=1}$, where $\boldsymbol{p_{t, k}} \in \mathbb{R}^{3\times 28\times 28}$. 
These patches are encoded using the pretrained vision encoder of Qwen-2.5-vl, producing embeddings that are dimensionally consistent with the text tokens, as depicted with blue inputs in \autoref{fig:method}. 
The vision encoder is frozen during the training stage.

\subsection{PIPE}
We propose \textbf{PIPE} to incorporate physical information into multimodal alignment for multimodal time series forecasting. Our proposed \textbf{PIPE} includes two cores: physics-informed positional indexing (\autoref{fig:method} \textcircled{\raisebox{-0.9pt}{1}}) and variant-frequency positional encoding (\autoref{fig:method} \textcircled{\raisebox{-0.9pt}{2}}). The algorithm can be found in \autoref{appendix:med}.

\subsubsection{Physics-Informed Positional Indexing}
We propose physics-informed positional indexing to integrate physical information into the model.
\paragraph{Schemes of indexing position IDs.}
Incorporating physical information into the model using positional IDs provides a direct solution without the need for additional structural complexity. We explore three indexing strategies to facilitate this integration:

(1) Sequential indexing. The most intuitive approach is to follow the standard transformer practice~\cite{vaswani2017attention} and ViT~\cite{dosovitskiy2020image}, using the sequence to index the position IDs.
In this scheme, the positional IDs are indexed linearly as:
\begin{equation}
\label{eq:sequence}
    position\_ids = [0, 1, 2, \ldots, seq\_len-1]
\end{equation}
where $seq\_len$ represents the total length of the input sequence, including both text tokens and vision tokens. This approach effectively encodes 1D sequential order but lacks explicit order information of the image (2D) or video (3D) for multimodal inputs.

(2) 3D indexing. 
Building on Qwen-2.5-VL~\cite{Qwen2.5-VL}, this method expands positional indexing to include three independent dimensions: temporal, height, and width, for the alignment of images and videos.
\begin{itemize}[leftmargin=*]
    \item Text tokens continue to use sequential indexing described in \autoref{eq:sequence}, while vision tokens are indexed based on their temporal and spatial attributes.
    \item Temporal positions of vision tokens are calculated as:
    \begin{equation}
        t = tokens\_per\_second \times temporal\_patch\_size / fps
    \end{equation}
    where $tokens\_per\_second$ dictates how many time steps are conceptually packed into a one-second interval of the video, $temporal\_patch\_size$ is the number of frames, and $fps$ is the video's frame rate.
    \item For spatial dimensions of vision tokens, the height and width positional IDs correspond to a patch grid ranging from $(0, 0)$ to $(N_{row}-1, N_{col}-1)$, where $N_{row}$ and $N_{col}$ are the numbers of image patching in height and width, respectively. Although this 3D indexing scheme aligns temporal and spatial order within vision tokens, it only captures the intra-relationship of positions with the input instance. It does not explicitly encode extra-physical properties such as time, latitude, and longitude, which are global knowledge among all instances in the dataset.
\end{itemize}
 
(3) Physics-Informed positional indexing (\autoref{fig:method}). To address the limitations of 3D indexing, we propose a novel physics-informed positional indexing scheme that explicitly integrates global knowledge of physical attributes into positional IDs.
\begin{itemize}[leftmargin=*]
    \item Text embeddings continue to use the sequence indexing scheme described in \autoref{eq:sequence}.
    \item Temporal positional IDs of vision tokens are computed based on the hourly progression of a given year. Specifically, the temporal position ID is calculated by:
    \begin{equation}
    \label{eq:t}
        t = t_{day} \times 24 + t_{hour}
    \end{equation}
    where $t_{day}$ is the day of the year (ranging from 0 to 365) and $t_{hour}$ represents the hour of the day (ranging from 0 to 23).
    This indexing introduces meaningful temporal patterns aligned with real-world time progression.
    \item The height and width positional IDs of vision tokens are determined using the latitude and longitude of the image patch centers.
\end{itemize}

To prevent the performance decreases caused by the conflicts between the physical information of vision tokens and the order information of text tokens (refer to the ablation experiment \autoref{sec:indexing}), we map the range of vision positional IDs ($t$: $0-8784$ ($8784$ hours in a year), $lat$: $0-180$, $lng$: $0-360$) to negative values. This avoids overlap with the text positional ID range, ensuring smooth multimodal integration. Moreover, temporal, latitudinal, and longitudinal dimensions are inherently independent, eliminating concerns about overlap.

After incorporating the cross-instance physical information among all input samples using physics-informed positional indexing, we apply RoPE~\cite{su2024roformer} on position IDs to encode intra-instance positional relationships within individual input samples.

\subsubsection{Variant-Frequency Positional Encoding}

We also merge the information of the physical variables into input embeddings.
To differentiate between physical variables, we modify the standard sinusoidal positional encoding (\autoref{eq:sin}) by introducing a variant-frequency sinusoidal function. 

\paragraph{Variant-frequency sinusoidal function} 
This function modifies the sine and cosine components and the target dimension based on the temporal, latitude, and longitude frequencies. \autoref{fig:method} illustrates the setting, \autoref{eq:detail_sin} in the Appendix gives the complete definition, and \autoref{fig:pe} visualizes the function. For conciseness, the function for image tokens can be formulated as:
\begin{align}
    PE_{(pos,2i)}=sin(\frac{pos}{10000^{2i/d_{model}}}\times\frac{2\pi}{p}) \\
    PE_{(pos,2i+1)}=cos(\frac{pos}{10000^{2i/d_{model}}}\times\frac{2\pi}{p}) \nonumber
\end{align}
where $pos$ can be $t_{day}$, $t_{hour}$, $lat$, and $lng$ depending on the dimensions. $t_{day}$ is the day of the year (ranging from 0 to 365) and $t_{hour}$ represents the hour of the day (ranging from 0 to 23). $lat$ is the latitude of the image token, and $lng$ is the longitude of the image token.
$p$ represents the wavelength specific to physics. For temporal data, $p_{day}=366$ and $p_{hour}=24$ and for spatial dimensions, $p_{latitude}=180$ and $p_{longitude}=360$. After the modification, the wavelengths form a geometric progression from $p$ to $p\cdot 10000 / 2$ for vision data.

Text tokens preserve the standard sinusoidal encoding to maintain compatibility with pretrained LLM structures.
These variant-frequency position encodings are added to the input embeddings at the bottom of the decoder stacks after they are divided by $d_{model}$. 
They map different physical variables to distinct frequency domains before incorporating them into the input embedding space.

\section{Experiments}
\label{sec:experment}
This section presents a systematic evaluation of the proposed method for the most representative multimodal time series forecasting task, typhoon forecasting. We first describe the datasets, baseline methods, and evaluation metrics, followed by the experiments and ablation studies.

\subsection{Dataset}
For multimodal time series forecasting, we utilize the open-source Digital Typhoon dataset~\cite{kitamoto2023digital}, the longest hourly satellite imagery collection dedicated to typhoon analysis spanning 40+ years (1978–2023) with a 5 $km$ spatial resolution. 
The spatial coverage of the dataset is the Western North Pacific basin.
The dataset includes 1,116 typhoon sequences and 192,956 images (resolution of 512×512 and resized to 224x224). The size of the dataset is different from the size in the original paper since the dataset is being regularly updated.

Typhoon track annotations, including intensity, latitude, and longitude, are sourced from the Best Track dataset~\cite{knapp2010international}. It is the best estimate, a globally recognized benchmark derived from retrospective post-event analysis. This metadata ensures reliable spatiotemporal grounding, as it synthesizes all available observational data to reconstruct each typhoon’s lifecycle with high precision. In our experiments, we will forecast three variables: intensity, latitude, and longitude. The dataset is split using a ratio of 0.7:0.15:0.15 based on the typhoon sequences as the original dataset.

\subsection{Baselines}
\paragraph{Domain models}
For typhoon forecasting, we compare our method against the state-of-the-art domain-specific NWP-based model: forecasting system of the European Centre for Medium-Range Weather Forecasts (ECMWF)~\cite{documentation2020part} and two environment-domain large models, Pangu~\cite{bi2023accurate} and GenCast~\cite{price2025probabilistic}, which serve as domain-specific benchmarks.
Additionally, we include comparisons with the domain practice method, Typhoon Intensity Forecasting based on the SHIPS method (TIFS)~\cite{ono2019operational}.
We report only the available performance from their paper and do not retrain the models, as we cannot reproduce these domain models.

\paragraph{AI models}
We train the state-of-the-art AI models with our dataset, including Transformer-based models (PatchTST~\cite{nie2022time}, iTransformer~\cite{liu2023itransformer}, Crossformer~\cite{wang2023crossformer}, TimeXer~\cite{wang2024timexer}) and linear-based models (TiDE~\cite{das2023long}), LLM-based model (One Fits ALL~\cite{zhou2023one}, AutoTimes~\cite{liu2024autotimes}), and other models (TimesNet~\cite{wu2022timesnet}, TimeMixer~\cite{wang2024timemixer}).
Due to their model design, they do not incorporate visual data. For the visual data integration, we include benchmark results reported in the original dataset publication (only the leading time of 12h is available)~\cite{kitamoto2023digital} and train the original Qwen-2.5-VL~\cite{Qwen2.5-VL}. 

\paragraph{Implementation Details}
Both our method and baselines use the same temporal settings with the same length of input and output sequences (12h).
For One Fits All and AutoTimes, we use their official implementations.
Other models without vision, their implementations are through the publicly available Time-Series-Library~\cite{wu2023timesnet}.
For the Qwen-2.5-VL model and \textbf{PIPE}, we use LLama-Factory~\cite{zheng2024llamafactory} for their implementation.
More implementation specifics, including hyperparameters and training protocols, are detailed in \autoref{appendix:implement}.

\subsection{Evaluation Metrics}
In NLP tasks, metrics such as ROUGE~\cite{lin2004rouge} and BLEU~\cite{papineni2002bleu} are commonly employed as the metrics.
In our cases, we focus on the numerical output. Specifically, for forecasting intensity, latitude, and longitude, we use Root Mean Square Error (RMSE) and Mean Absolute Error (MAE) as primary metrics. 
When the model accurately predicts these numerical values based on satellite images, we consider it to have effectively aligned the satellite imagery with the time series data.
Additionally, we use geographiclib~\cite{karney2013algorithms} to calculate the position error of typhoon tracks based on the latitude and longitude, following the domain practice.

\subsection{Main Results}
The forecasting performance of multimodal time series models is summarized in \autoref{tab:6h} (6-hour lead time) and \autoref{tab:12h} (12-hour lead time), with the best results highlighted in \textbf{bold} and the second-best results highlighted in \underline{underline}.

Overall, our method achieves state-of-the-art performance across the majority of evaluation metrics, demonstrating the efficacy of the proposed \textbf{PIPE} in integrating physical information during multimodal alignment.
For the 6-hour lead time (\autoref{tab:6h}), our model outperforms baselines in most metrics. 
For example, it shows 12\% improvement of MAE for typhoon intensity forecasting when compared to the best w/o vision models TiDE.
The sole exception is the RMSE for intensity forecasting, where TiDE and PatchTST exhibit marginally superior performance. 
These results show the effectiveness of our approach. 
A critical observation is the consistent superiority of models incorporating vision data over unimodal alternatives. This finding emphasizes the importance of leveraging multimodal inputs to enhance forecasting accuracy in complex spatiotemporal tasks. 

\begin{table*}[t]
\caption{Multimodal time series forecasting results (leading time is 6h).}
\label{tab:6h}
\centering
\small
\resizebox{\textwidth}{!}{

\begin{tabular}{p{0.3cm}p{3.4cm}|cc|cc|cc|c} 
\toprule
& \multirow{2}{*}{\textbf{Models}}& \multicolumn{2}{c|}{Intensity (hPa)} &\multicolumn{2}{c|}{Latitude ($^\circ$)} &\multicolumn{2}{c|}{Longitude ($^\circ$)} & Distance (km)\\ 
& &MAE &RMSE &MAE &RMSE &MAE &RMSE & MAE\\

\midrule
\parbox[t]{3mm}{\multirow{4}{*}{\rotatebox[origin=c]{90}{domain}}} & ECMWF-HRES~\cite{documentation2020part}  & \multicolumn{2}{c|}{\multirow{3}{*}{$\setminus$}} & \multicolumn{2}{c|}{\multirow{4}{*}{$\setminus$}} & \multicolumn{2}{c|}{\multirow{4}{*}{$\setminus$}} & 27.181  \\
& PanGu~\cite{bi2023accurate}  & \multicolumn{2}{c|}{} & \multicolumn{2}{c|}{} & \multicolumn{2}{c|}{} & 32.892  \\
& GenCast~\cite{price2025probabilistic}  & \multicolumn{2}{c|}{} & \multicolumn{2}{c|}{} & \multicolumn{2}{c|}{} & 20.331  \\ 
& TIFS~\cite{ono2019operational} & $\setminus$ & 7.292 & \multicolumn{2}{c|}{} & \multicolumn{2}{c|}{} & $\setminus$  \\ 
\midrule
\parbox[t]{3mm}{\multirow{9}{*}{\rotatebox[origin=c]{90}{w/o vision}}} & PatchTST~\cite{nie2022time} & 1.806 & \underline{2.867} & 0.199 & 0.266 & 0.322 & 0.404 & 44.537 \\
 & iTransformer~\cite{liu2023itransformer} & 1.848 & 2.979 & 0.164 & 0.231 & 0.203 & 0.281 & 31.248 \\
 & Crossformer~\cite{wang2023crossformer} & 2.389 & 3.599 & 0.310 & 0.418 & 0.520 & 0.684 & 71.216 \\
 & TimeXer~\cite{wang2024timexer}) & 3.037 & 4.523 & 0.306 & 0.411 & 0.411 & 0.538 & 59.720 \\
 & TiDE~\cite{das2023long} & 1.724 & \textbf{2.819} & 0.161 & 0.224 & 0.237 & 0.312 & 34.068 \\
 & One Fits All~\cite{zhou2023one}) & 1.849 & 2.976 & 0.170 & 0.239 & 0.211 & 0.290 & 32.450 \\
 & AutoTimes~\cite{liu2024autotimes} & 1.991 & 3.088 & 0.190 & 0.265 & 0.279 & 0.364 & 40.036 \\ 
 & TimesNet~\cite{wu2022timesnet} & 2.401 & 3.711 & 0.465 & 0.630 & 0.855 & 1.124 & 113.718 \\
 & TimeMixer~\cite{wang2024timemixer} & 1.913 & 2.973 & 0.177 & 0.237 & 0.238 & 0.313 & 35.374 \\
\midrule
\parbox[t]{3mm}{\multirow{3}{*}{\rotatebox[origin=l]{90}{vision}}} & Qwen-2.5-VL-3B~\cite{Qwen2.5-VL} & \underline{1.617} & 3.231 & \underline{0.087} & \underline{0.162} & \underline{0.103} & \underline{0.187} & \underline{17.129} \\ 
\cmidrule(l{2em}r{0em}){1-9}
 & \textbf{PIPE-3B} & \textbf{1.515} & 2.981 & \textbf{0.084} & \textbf{0.159} & \textbf{0.095} & \textbf{0.178} & \textbf{16.275} \\
\bottomrule
\end{tabular}

}
\end{table*}

\subsection{Regression Analysis}

\begin{figure}[t]
  \centering
  \includegraphics[width=\linewidth]{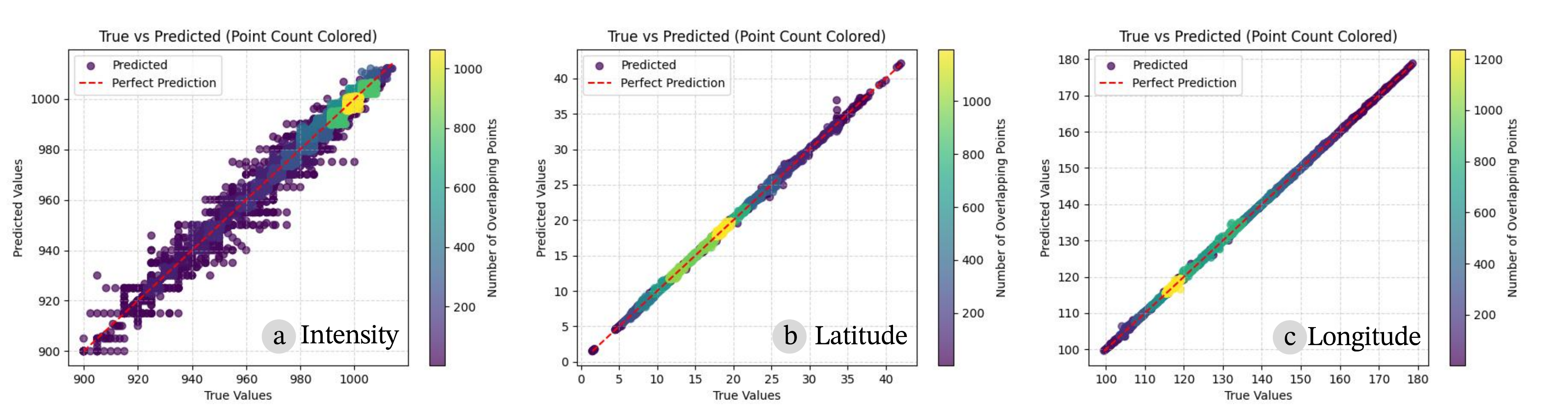}
  \caption{The visualization for the regression results between predicted values and true values (leading time is 6 hours). (a) Plots for intensity. (b) Plots for latitude. (c) Plots for longitude.}
  \label{fig:regression-6h}
\end{figure}

The regression results of all test typhoon sequences (\autoref{fig:regression-6h}) demonstrate that our method achieves accurate typhoon predictions. Notably, the model exhibits superior performance in location forecasting compared to intensity forecasting, which may be attributed to the richer spatial information provided by satellite imagery for tracking movement. Additionally, the model shows better predictive performance when typhoon intensity is weaker (i.e., higher central pressure, around $1000\;hpa$).

\subsection{Ablation Study}
\label{sec:ablation}
The results of the ablation study are presented in \autoref{tab:ablation_6h} and \autoref{tab:ablation_12h}, with the best performance highlighted in \textbf{bold}.
We systematically evaluate three critical components: vision inclusion, physics-informed position indexing, and variant-frequency sinusoidal function.

\paragraph{The gain of aligning vision}
The inclusion of satellite vision data yields significant improvements in forecasting accuracy.
Specifically, the MAE for intensity forecasting improves by up to 8\% when the leading time is set to 6 hours.
This demonstrates that cross-modal learning effectively leverages spatial patterns in satellite imagery to complement time series data.

\paragraph{Comparison of schemes of indexing position IDs}
\label{sec:indexing}
Our physics-informed indexing scheme addresses the critical challenge of preserving physical knowledge while avoiding token order conflicts.
To assess its effectiveness, we compare different schemes for indexing position IDs.
Specifically, we evaluate the performance by (a) removing the 3D indexing scheme (replacing it with sequential indexing), and (b) removing physics-informed indexing while retaining the 3D indexing scheme.
The results show that while sequential indexing and 3D indexing perform similarly, both exhibit a noticeable performance degradation (6\% for MAE of intensity forecasting) compared to the physics-informed indexing scheme. Avoiding the overlap between the physical information of vision tokens and the order information of text tokens is critical. There is a dramatic performance decrease when they share overlapping ranges (e.g., longitude: $0-360$ and text tokens: $0-seq_{len}$ ($seq_{len}$ is the number of text tokens)).
By mapping the position IDs of vision tokens to negative values, we preserve the physical information and resolve such conflicts, leading to improved performance.

\paragraph{The gain of integrating physical variables' frequency}
The incorporation of frequency characteristics of physical variables improves physical variable modeling.
We show the importance by removing the entire sinusoidal function and only removing the variant-frequency sinusoidal function. The results reveal that our designed sinusoidal function plays a crucial role in aligning the model with the frequency information of physical variables. Its inclusion enhances the model's ability to leverage these variables effectively, leading to improved performance.

Every component contributes to the multimodal time series forecasting, with vision alignment providing complementary visual patterns, the physics-informed indexing scheme ensuring physical knowledge integration, and the variant-frequency sinusoidal function incorporating physical variables' frequency information.

\begin{table*}[t]
\caption{The results of the ablation study (leading time is 6h).}
\label{tab:ablation_6h}
\centering
\resizebox{\textwidth}{!}{
\begin{tabular}{l|cc|cc|cc|c} 
\toprule
\multirow{2}{*}{\textbf{Models}}& \multicolumn{2}{c|}{Intensity (hPa)} &\multicolumn{2}{c|}{Latitude ($^\circ$)} &\multicolumn{2}{c|}{Longitude ($^\circ$)} & Distance (km)\\ 
&MAE &RMSE &MAE &RMSE &MAE &RMSE & MAE\\

\midrule
w/o vision & 1.646 & 3.220 & 0.088 & 0.160 & 0.102 & 0.193 & 17.235 \\
\midrule
w/o 3D indexing (using sequence) & 1.628 & 3.749 & 0.087 & 0.163 & 0.102 & 0.185 & 17.084 \\
w/o physics-informed indexing (using 3D) & 1.617 & 3.231 & 0.087 & 0.162 & 0.103 & 0.187 & 17.129 \\ 
w/o negative indexing & 1.961 & 3.926 & 0.206 & 0.360 & 0.388 & 0.674 & 53.548 \\
\midrule
w/o entire sinusoidal function & 1.545 & 3.053 & 0.085 & \textbf{0.157} & 0.097 & 0.180 & 16.554 \\
w/o variant-frequency sinusoidal function & 1.639 & 3.178 & 0.086 & 0.161 & 0.101 & 0.604 & 16.860 \\

\midrule
\textbf{PIPE-3B} & \textbf{1.515} & \textbf{2.981} & \textbf{0.084} & 0.159 & \textbf{0.095} & \textbf{0.178} & \textbf{16.275} \\
\bottomrule
\end{tabular}
}
\end{table*}

\section{Conclusion}

This paper proposes a multimodal time series forecasting task and addresses the challenge brought by integrating satellite imagery. 
Existing approaches only focus on pixel-level features, overlooking the rich temporal and geophysical context embedded within vision data.
We propose \textbf{p}hysics-\textbf{i}nformed \textbf{p}osition \textbf{e}ncoding (\textbf{PIPE}). 
Experimental results demonstrate that \textbf{PIPE} achieves state-of-the-art performance across multiple benchmarks. Ablation studies further validate the distinct contributions of each component. Future work will explore the integration of additional physical domain knowledge, such as physical laws and constraints, to enhance real-world applicability.

\bibliography{references}

\begin{thebibliography}{10}

\bibitem{abnar2020quantifying}
Samira Abnar and Willem Zuidema.
\newblock Quantifying attention flow in transformers.
\newblock In Dan Jurafsky, Joyce Chai, Natalie Schluter, and Joel Tetreault, editors, {\em Proceedings of the 58th Annual Meeting of the Association for Computational Linguistics}, pages 4190--4197, Online, July 2020. Association for Computational Linguistics.

\bibitem{Qwen2.5-VL}
Shuai Bai, Keqin Chen, Xuejing Liu, Jialin Wang, Wenbin Ge, Sibo Song, Kai Dang, Peng Wang, Shijie Wang, Jun Tang, Humen Zhong, Yuanzhi Zhu, Mingkun Yang, Zhaohai Li, Jianqiang Wan, Pengfei Wang, Wei Ding, Zheren Fu, Yiheng Xu, Jiabo Ye, Xi~Zhang, Tianbao Xie, Zesen Cheng, Hang Zhang, Zhibo Yang, Haiyang Xu, and Junyang Lin.
\newblock Qwen2.5-vl technical report.
\newblock {\em arXiv preprint arXiv:2502.13923}, 2025.

\bibitem{bi2023accurate}
Kaifeng Bi, Lingxi Xie, Hengheng Zhang, Xin Chen, Xiaotao Gu, and Qi~Tian.
\newblock Accurate medium-range global weather forecasting with 3d neural networks.
\newblock {\em Nature}, 619(7970):533--538, 2023.

\bibitem{chang2025llm4ts}
Ching Chang, Wei-Yao Wang, Wen-Chih Peng, and Tien-Fu Chen.
\newblock Llm4ts: Aligning pre-trained llms as data-efficient time-series forecasters.
\newblock {\em ACM Trans. Intell. Syst. Technol.}, 16(3), April 2025.

\bibitem{chen2024pathformer}
Peng Chen, Yingying Zhang, Yunyao Cheng, Yang Shu, Yihang Wang, Qingsong Wen, Bin Yang, and Chenjuan Guo.
\newblock Pathformer: Multi-scale transformers with adaptive pathways for time series forecasting.
\newblock {\em arXiv preprint arXiv:2402.05956}, 2024.

\bibitem{dagdelen2024structured}
John Dagdelen, Alexander Dunn, Sanghoon Lee, Nicholas Walker, Andrew~S Rosen, Gerbrand Ceder, Kristin~A Persson, and Anubhav Jain.
\newblock Structured information extraction from scientific text with large language models.
\newblock {\em Nature Communications}, 15(1):1418, 2024.

\bibitem{dai2019transformer}
Zihang Dai, Zhilin Yang, Yiming Yang, Jaime Carbonell, Quoc Le, and Ruslan Salakhutdinov.
\newblock Transformer-{XL}: Attentive language models beyond a fixed-length context.
\newblock In Anna Korhonen, David Traum, and Llu{\'i}s M{\`a}rquez, editors, {\em Proceedings of the 57th Annual Meeting of the Association for Computational Linguistics}, pages 2978--2988, Florence, Italy, July 2019. Association for Computational Linguistics.

\bibitem{das2023long}
Abhimanyu Das, Weihao Kong, Andrew Leach, Shaan Mathur, Rajat Sen, and Rose Yu.
\newblock Long-term forecasting with tide: Time-series dense encoder.
\newblock {\em arXiv preprint arXiv:2304.08424}, 2023.

\bibitem{documentation2020part}
IFS DOCUMENTATION-Cy40r1.
\newblock Part v: Ensemble prediction system.
\newblock 2020.

\bibitem{dosovitskiy2020image}
Alexey Dosovitskiy, Lucas Beyer, Alexander Kolesnikov, Dirk Weissenborn, Xiaohua Zhai, Thomas Unterthiner, Mostafa Dehghani, Matthias Minderer, Georg Heigold, Sylvain Gelly, et~al.
\newblock An image is worth 16x16 words: Transformers for image recognition at scale.
\newblock {\em arXiv preprint arXiv:2010.11929}, 2020.

\bibitem{ho1998use}
Siu~Lau Ho and Min Xie.
\newblock The use of arima models for reliability forecasting and analysis.
\newblock {\em Computers \& industrial engineering}, 35(1-2):213--216, 1998.

\bibitem{hochreiter1997long}
Sepp Hochreiter and J{\"u}rgen Schmidhuber.
\newblock Long short-term memory.
\newblock {\em Neural computation}, 9(8):1735--1780, 1997.

\bibitem{hu2022lora}
Edward~J Hu, Yelong Shen, Phillip Wallis, Zeyuan Allen-Zhu, Yuanzhi Li, Shean Wang, Lu~Wang, Weizhu Chen, et~al.
\newblock Lora: Low-rank adaptation of large language models.
\newblock {\em ICLR}, 1(2):3, 2022.

\bibitem{jean2016combining}
Neal Jean, Marshall Burke, Michael Xie, W~Matthew Alampay~Davis, David~B Lobell, and Stefano Ermon.
\newblock Combining satellite imagery and machine learning to predict poverty.
\newblock {\em Science}, 353(6301):790--794, 2016.

\bibitem{jia2024gpt4mts}
Furong Jia, Kevin Wang, Yixiang Zheng, Defu Cao, and Yan Liu.
\newblock Gpt4mts: Prompt-based large language model for multimodal time-series forecasting.
\newblock In {\em Proceedings of the AAAI Conference on Artificial Intelligence}, volume~38, pages 23343--23351, 2024.

\bibitem{jin2023time}
Ming Jin, Shiyu Wang, Lintao Ma, Zhixuan Chu, James~Y Zhang, Xiaoming Shi, Pin-Yu Chen, Yuxuan Liang, Yuan-Fang Li, Shirui Pan, and Qingsong Wen.
\newblock {Time-LLM}: Time series forecasting by reprogramming large language models.
\newblock In {\em International Conference on Learning Representations (ICLR)}, 2024.

\bibitem{karney2013algorithms}
Charles~FF Karney.
\newblock Algorithms for geodesics.
\newblock {\em Journal of Geodesy}, 87:43--55, 2013.

\bibitem{ke2021rethinking}
Guolin Ke, Di~He, and Tie-Yan Liu.
\newblock Rethinking positional encoding in language pre-training.
\newblock In {\em International Conference on Learning Representations}, 2021.

\bibitem{kitaev2020reformer}
Nikita Kitaev, Lukasz Kaiser, and Anselm Levskaya.
\newblock Reformer: The efficient transformer.
\newblock In {\em International Conference on Learning Representations}, 2020.

\bibitem{kitamoto2023digital}
Asanobu Kitamoto, Jared Hwang, Bastien Vuillod, Lucas Gautier, Yingtao Tian, and Tarin Clanuwat.
\newblock Digital typhoon: Long-term satellite image dataset for the spatio-temporal modeling of tropical cyclones.
\newblock {\em Advances in Neural Information Processing Systems}, 36:40623--40636, 2023.

\bibitem{knapp2010international}
Kenneth~R Knapp, Michael~C Kruk, David~H Levinson, Howard~J Diamond, and Charles~J Neumann.
\newblock The international best track archive for climate stewardship (ibtracs) unifying tropical cyclone data.
\newblock {\em Bulletin of the American Meteorological Society}, 91(3):363--376, 2010.

\bibitem{li2023blip}
Junnan Li, Dongxu Li, Silvio Savarese, and Steven Hoi.
\newblock Blip-2: Bootstrapping language-image pre-training with frozen image encoders and large language models.
\newblock In {\em International conference on machine learning}, pages 19730--19742. PMLR, 2023.

\bibitem{li2024urbangpt}
Zhonghang Li, Lianghao Xia, Jiabin Tang, Yong Xu, Lei Shi, Long Xia, Dawei Yin, and Chao Huang.
\newblock Urbangpt: Spatio-temporal large language models.
\newblock In {\em Proceedings of the 30th ACM SIGKDD Conference on Knowledge Discovery and Data Mining}, pages 5351--5362, 2024.

\bibitem{lin2004rouge}
Chin-Yew Lin.
\newblock {ROUGE}: A package for automatic evaluation of summaries.
\newblock In {\em Text Summarization Branches Out}, pages 74--81, Barcelona, Spain, July 2004. Association for Computational Linguistics.

\bibitem{lin2022cat}
Hezheng Lin, Xing Cheng, Xiangyu Wu, and Dong Shen.
\newblock Cat: Cross attention in vision transformer.
\newblock In {\em 2022 IEEE international conference on multimedia and expo (ICME)}, pages 1--6. IEEE, 2022.

\bibitem{liu2023visual}
Haotian Liu, Chunyuan Li, Qingyang Wu, and Yong~Jae Lee.
\newblock Visual instruction tuning.
\newblock {\em Advances in neural information processing systems}, 36:34892--34916, 2023.

\bibitem{liu2024unitime}
Xu~Liu, Junfeng Hu, Yuan Li, Shizhe Diao, Yuxuan Liang, Bryan Hooi, and Roger Zimmermann.
\newblock Unitime: A language-empowered unified model for cross-domain time series forecasting.
\newblock In {\em Proceedings of the ACM Web Conference 2024}, pages 4095--4106, 2024.

\bibitem{liu2020improving}
Xuanqing Liu, Hsiang-Fu Yu, Inderjit Dhillon, and Cho-Jui Hsieh.
\newblock Learning to encode position for transformer with continuous dynamical model.
\newblock In Hal~Daumé III and Aarti Singh, editors, {\em Proceedings of the 37th International Conference on Machine Learning}, volume 119 of {\em Proceedings of Machine Learning Research}, pages 6327--6335. PMLR, 13--18 Jul 2020.

\bibitem{liu2023itransformer}
Yong Liu, Tengge Hu, Haoran Zhang, Haixu Wu, Shiyu Wang, Lintao Ma, and Mingsheng Long.
\newblock itransformer: Inverted transformers are effective for time series forecasting.
\newblock {\em arXiv preprint arXiv:2310.06625}, 2023.

\bibitem{liu2024autotimes}
Yong Liu, Guo Qin, Xiangdong Huang, Jianmin Wang, and Mingsheng Long.
\newblock Autotimes: Autoregressive time series forecasters via large language models.
\newblock {\em arXiv preprint arXiv:2402.02370}, 2024.

\bibitem{loshchilov2017decoupled}
Ilya Loshchilov and Frank Hutter.
\newblock Decoupled weight decay regularization.
\newblock {\em arXiv preprint arXiv:1711.05101}, 2017.

\bibitem{nie2022time}
Yuqi Nie, Nam~H Nguyen, Phanwadee Sinthong, and Jayant Kalagnanam.
\newblock A time series is worth 64 words: Long-term forecasting with transformers.
\newblock {\em arXiv preprint arXiv:2211.14730}, 2022.

\bibitem{ono2019operational}
Marika Ono, Shoji Notsuhara, Junya Fukuda, Yohko Igarashi, and Kotaro Bessho.
\newblock Operational use of the typhoon intensity forecasting scheme based on ships (tifs) and commencement of five-day tropical cyclone intensity forecasts.
\newblock {\em ENE}, 128(40):128--7, 2019.

\bibitem{openai2024gpt4technicalreport}
OpenAI et~al.
\newblock Gpt-4 technical report, 2024.

\bibitem{papineni2002bleu}
Kishore Papineni, Salim Roukos, Todd Ward, and Wei-Jing Zhu.
\newblock Bleu: a method for automatic evaluation of machine translation.
\newblock In {\em Proceedings of the 40th annual meeting of the Association for Computational Linguistics}, pages 311--318, 2002.

\bibitem{paszke2019pytorch}
Adam Paszke, Sam Gross, Francisco Massa, Adam Lerer, James Bradbury, Gregory Chanan, Trevor Killeen, Zeming Lin, Natalia Gimelshein, Luca Antiga, Alban Desmaison, Andreas Kopf, Edward Yang, Zachary DeVito, Martin Raison, Alykhan Tejani, Sasank Chilamkurthy, Benoit Steiner, Lu~Fang, Junjie Bai, and Soumith Chintala.
\newblock Pytorch: An imperative style, high-performance deep learning library.
\newblock In H.~Wallach, H.~Larochelle, A.~Beygelzimer, F.~d\textquotesingle Alch\'{e}-Buc, E.~Fox, and R.~Garnett, editors, {\em Advances in Neural Information Processing Systems}, volume~32. Curran Associates, Inc., 2019.

\bibitem{press2021shortformer}
Ofir Press, Noah~A. Smith, and Mike Lewis.
\newblock Shortformer: Better language modeling using shorter inputs.
\newblock In Chengqing Zong, Fei Xia, Wenjie Li, and Roberto Navigli, editors, {\em Proceedings of the 59th Annual Meeting of the Association for Computational Linguistics and the 11th International Joint Conference on Natural Language Processing (Volume 1: Long Papers)}, pages 5493--5505, Online, August 2021. Association for Computational Linguistics.

\bibitem{price2025probabilistic}
Ilan Price, Alvaro Sanchez-Gonzalez, Ferran Alet, Tom~R Andersson, Andrew El-Kadi, Dominic Masters, Timo Ewalds, Jacklynn Stott, Shakir Mohamed, Peter Battaglia, et~al.
\newblock Probabilistic weather forecasting with machine learning.
\newblock {\em Nature}, 637(8044):84--90, 2025.

\bibitem{radford2021learning}
Alec Radford, Jong~Wook Kim, Chris Hallacy, Aditya Ramesh, Gabriel Goh, Sandhini Agarwal, Girish Sastry, Amanda Askell, Pamela Mishkin, Jack Clark, et~al.
\newblock Learning transferable visual models from natural language supervision.
\newblock In {\em International conference on machine learning}, pages 8748--8763. PmLR, 2021.

\bibitem{rumelhart1986learning}
David~E Rumelhart, Geoffrey~E Hinton, and Ronald~J Williams.
\newblock Learning representations by back-propagating errors.
\newblock {\em nature}, 323(6088):533--536, 1986.

\bibitem{shaw2018self}
Peter Shaw, Jakob Uszkoreit, and Ashish Vaswani.
\newblock Self-attention with relative position representations.
\newblock In Marilyn Walker, Heng Ji, and Amanda Stent, editors, {\em Proceedings of the 2018 Conference of the North {A}merican Chapter of the Association for Computational Linguistics: Human Language Technologies, Volume 2 (Short Papers)}, pages 464--468, New Orleans, Louisiana, June 2018. Association for Computational Linguistics.

\bibitem{su2024roformer}
Jianlin Su, Murtadha Ahmed, Yu~Lu, Shengfeng Pan, Wen Bo, and Yunfeng Liu.
\newblock Roformer: Enhanced transformer with rotary position embedding.
\newblock {\em Neurocomput.}, 568(C), February 2024.

\bibitem{sun2024test}
Chenxi Sun, Hongyan Li, Yaliang Li, and Shenda Hong.
\newblock {TEST}: Text prototype aligned embedding to activate {LLM}'s ability for time series.
\newblock In {\em The Twelfth International Conference on Learning Representations}, 2024.

\bibitem{team2023gemini}
Gemini Team, Rohan Anil, Sebastian Borgeaud, Yonghui Wu, Jean-Baptiste Alayrac, Jiahui Yu, Radu Soricut, Johan Schalkwyk, Andrew~M Dai, Anja Hauth, et~al.
\newblock Gemini: a family of highly capable multimodal models.
\newblock {\em arXiv preprint arXiv:2312.11805}, 2023.

\bibitem{vaswani2017attention}
Ashish Vaswani, Noam Shazeer, Niki Parmar, Jakob Uszkoreit, Llion Jones, Aidan~N Gomez, {\L}ukasz Kaiser, and Illia Polosukhin.
\newblock Attention is all you need.
\newblock {\em Advances in neural information processing systems}, 30, 2017.

\bibitem{veillette2020sevir}
Mark Veillette, Siddharth Samsi, and Chris Mattioli.
\newblock Sevir: A storm event imagery dataset for deep learning applications in radar and satellite meteorology.
\newblock {\em Advances in Neural Information Processing Systems}, 33:22009--22019, 2020.

\bibitem{wang2025chattime}
Chengsen Wang, Qi~Qi, Jingyu Wang, Haifeng Sun, Zirui Zhuang, Jinming Wu, Lei Zhang, and Jianxin Liao.
\newblock Chattime: A unified multimodal time series foundation model bridging numerical and textual data.
\newblock In {\em Proceedings of the AAAI Conference on Artificial Intelligence}, volume~39, pages 12694--12702, 2025.

\bibitem{wang2024timemixer}
Shiyu Wang, Haixu Wu, Xiaoming Shi, Tengge Hu, Huakun Luo, Lintao Ma, James~Y Zhang, and Jun Zhou.
\newblock Timemixer: Decomposable multiscale mixing for time series forecasting.
\newblock {\em arXiv preprint arXiv:2405.14616}, 2024.

\bibitem{wang2023crossformer}
Wenxiao Wang, Wei Chen, Qibo Qiu, Long Chen, Boxi Wu, Binbin Lin, Xiaofei He, and Wei Liu.
\newblock Crossformer++: A versatile vision transformer hinging on cross-scale attention.
\newblock {\em IEEE Transactions on Pattern Analysis and Machine Intelligence}, 46(5):3123--3136, 2023.

\bibitem{wang2023contrast}
Yihe Wang, Yu~Han, Haishuai Wang, and Xiang Zhang.
\newblock Contrast everything: A hierarchical contrastive framework for medical time-series.
\newblock {\em Advances in Neural Information Processing Systems}, 36:55694--55717, 2023.

\bibitem{wang2024timexer}
Yuxuan Wang, Haixu Wu, Jiaxiang Dong, Guo Qin, Haoran Zhang, Yong Liu, Yunzhong Qiu, Jianmin Wang, and Mingsheng Long.
\newblock Timexer: Empowering transformers for time series forecasting with exogenous variables.
\newblock {\em arXiv preprint arXiv:2402.19072}, 2024.

\bibitem{wu2021da}
Chuhan Wu, Fangzhao Wu, and Yongfeng Huang.
\newblock {DA}-transformer: Distance-aware transformer.
\newblock In Kristina Toutanova, Anna Rumshisky, Luke Zettlemoyer, Dilek Hakkani-Tur, Iz~Beltagy, Steven Bethard, Ryan Cotterell, Tanmoy Chakraborty, and Yichao Zhou, editors, {\em Proceedings of the 2021 Conference of the North American Chapter of the Association for Computational Linguistics: Human Language Technologies}, pages 2059--2068, Online, June 2021. Association for Computational Linguistics.

\bibitem{wu2022timesnet}
Haixu Wu, Tengge Hu, Yong Liu, Hang Zhou, Jianmin Wang, and Mingsheng Long.
\newblock Timesnet: Temporal 2d-variation modeling for general time series analysis.
\newblock {\em arXiv preprint arXiv:2210.02186}, 2022.

\bibitem{wu2023timesnet}
Haixu Wu, Tengge Hu, Yong Liu, Hang Zhou, Jianmin Wang, and Mingsheng Long.
\newblock Timesnet: Temporal 2d-variation modeling for general time series analysis.
\newblock In {\em International Conference on Learning Representations}, 2023.

\bibitem{wu2021autoformer}
Haixu Wu, Jiehui Xu, Jianmin Wang, and Mingsheng Long.
\newblock Autoformer: Decomposition transformers with auto-correlation for long-term series forecasting.
\newblock {\em Advances in neural information processing systems}, 34:22419--22430, 2021.

\bibitem{yuan2020self}
Yuan Yuan and Lei Lin.
\newblock Self-supervised pretraining of transformers for satellite image time series classification.
\newblock {\em IEEE Journal of Selected Topics in Applied Earth Observations and Remote Sensing}, 14:474--487, 2020.

\bibitem{zhang2024sageformer}
Zhenwei Zhang, Linghang Meng, and Yuantao Gu.
\newblock Sageformer: Series-aware framework for long-term multivariate time-series forecasting.
\newblock {\em IEEE Internet of Things Journal}, 11(10):18435--18448, 2024.

\bibitem{zheng2024llamafactory}
Yaowei Zheng, Richong Zhang, Junhao Zhang, Yanhan Ye, Zheyan Luo, Zhangchi Feng, and Yongqiang Ma.
\newblock Llamafactory: Unified efficient fine-tuning of 100+ language models.
\newblock In {\em Proceedings of the 62nd Annual Meeting of the Association for Computational Linguistics (Volume 3: System Demonstrations)}, Bangkok, Thailand, 2024. Association for Computational Linguistics.

\bibitem{zhou2021informer}
Haoyi Zhou, Shanghang Zhang, Jieqi Peng, Shuai Zhang, Jianxin Li, Hui Xiong, and Wancai Zhang.
\newblock Informer: Beyond efficient transformer for long sequence time-series forecasting.
\newblock In {\em Proceedings of the AAAI conference on artificial intelligence}, volume~35, pages 11106--11115, 2021.

\bibitem{zhou2023one}
Tian Zhou, Peisong Niu, Liang Sun, Rong Jin, et~al.
\newblock One fits all: Power general time series analysis by pretrained lm.
\newblock {\em Advances in neural information processing systems}, 36:43322--43355, 2023.

\end{thebibliography}




\appendix


\section{Prompts Design}
\label{appendix:prompt}
We show our prompt design for the multimodal time series forecasting, taking one instance of Typhoon Yutu as an example.
\tcbset{width=\textwidth, colframe=black}
\begin{tcolorbox}[title = {System Prompt \& Task Instruction}]
You are a typhoon forecasting expert. Below are the past 12 hours of typhoon data and the corresponding satellite images. Your task is to forecast the hourly data of the typhoon for the next 12 hours, providing the forecast latitude, longitude, pressure in the same format as the past data format. 
\end{tcolorbox}

\tcbset{width=\textwidth, colframe=black}
\begin{tcolorbox}[title = {Past Data}]
The corresponding satellite images are: <image> <image> <image> <image> <image> <image> <image> <image> <image> <image> <image> <image>.  The historical hourly data from 2018-10-23 01:00:00 to 2018-10-23 12:00:00 is \{latitude: [11.65, 11.7, 11.75, 11.8, 11.85, 11.9, 11.95, 11.99, 12.04, 12.09, 12.14, 12.2], longitude: [151.61, 151.41, 151.2, 150.99, 150.79, 150.6, 150.42, 150.26, 150.11, 149.97, 149.83, 149.7], pressure: [974.2, 973.3, 972.5, 971.7, 970.8, 970.0, 967.5, 965.0, 962.5, 960.0, 957.5, 955.0]\}.
\end{tcolorbox}

\tcbset{width=\textwidth, colframe=black}
\begin{tcolorbox}[title = {Label Data}]
The forecast hourly data is: \{latitude: [12.26, 12.34, 12.42, 12.5, 12.6, 12.7, 12.81, 12.93, 13.05, 13.17, 13.29, 13.4], longitude: [149.57, 149.44, 149.31, 149.18, 149.04, 148.9, 148.76, 148.61, 148.46, 148.31, 148.15, 148.0], pressure: [954.2, 953.3, 952.5, 951.7, 950.8, 950.0, 945.8, 941.7, 937.5, 933.3, 929.2, 925.0].\}
\end{tcolorbox}

\tcbset{width=\textwidth, colframe=black}
\begin{tcolorbox}[title = {Image Data}]
\includegraphics[width=\linewidth,keepaspectratio]{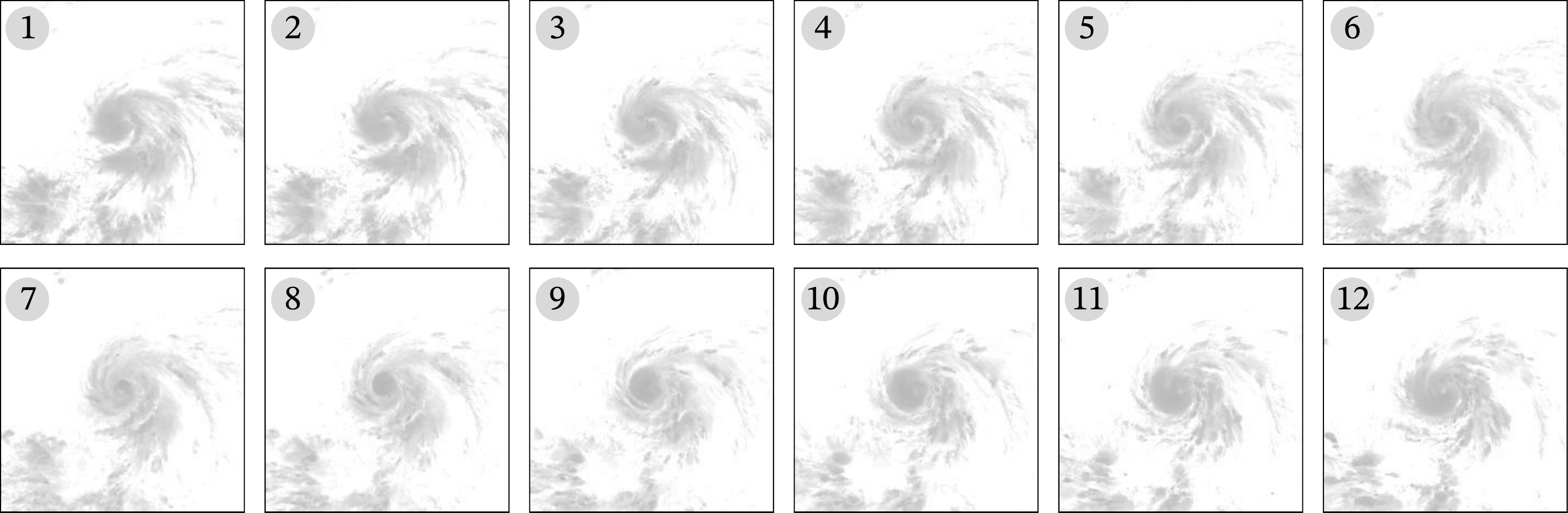}
\end{tcolorbox}

\section{Method}
\label{appendix:med}
\subsection{Algorithms}

In this section, we present the algorithms for PIPE (\autoref{alg:pipe}), which integrates physical information into VLMs.
To begin, we extract the required physical information (\autoref{alg:extract}) from the time series data corresponding to each satellite image. This information includes the timestamp and geocoordinates for the typhoon's eye. Additionally, since satellite images are divided into patches, we calculate the geocoordinates for the center of each patch.
Next, we incorporate this physical information into the positional encoding of VLMs through physics-informed positional indexing (\autoref{alg:position_indexing}). Beyond indexing, we adapt the sinusoidal function by introducing variant-frequency sinusoidal encoding (\autoref{alg:vf_sin}), which embeds the frequency attributes of the variables into the positional embedding. This enhanced positional embedding is then added to the input embedding of the corresponding image tokens.
Finally, these integrations are utilized to predict the next token, enabling the model to leverage both spatial and physical context effectively.

\begin{algorithm}[H]
\caption{PIPE}
\label{alg:pipe}
\begin{algorithmic}[1]
\Require{Time series input $\boldsymbol{x}$, corresponding image $\boldsymbol{i}$}
\Ensure{Next token prediction $\boldsymbol{T}_{next}$}
\State $\boldsymbol{T}_{text}, \boldsymbol{T}_{image}$ = tokenizer($\boldsymbol{x}$), vision\_encoder($\boldsymbol{i}$) \Comment{Tokenization}
\State $\boldsymbol{t}, \boldsymbol{lat}, \boldsymbol{lng} \gets \text{get\_physic}(\boldsymbol{x}, \boldsymbol{T}_{image})$ \Comment{Extract physical info (\autoref{alg:extract}): $\boldsymbol{t}, \boldsymbol{lat}, \boldsymbol{lng} \in \mathbb{R}^{1\times len(\boldsymbol{T}_{image})}$}
\State $\boldsymbol{ids} \gets \text{position\_indexing}(\boldsymbol{T}_{text}, \boldsymbol{T}_{image}, \boldsymbol{t}, \boldsymbol{lat}, \boldsymbol{lng})$  \Comment{Compute physics-informed indices (\autoref{alg:position_indexing}): $\boldsymbol{ids} \in \mathbb{R}^{3\times len(\boldsymbol{T}_{text} + \boldsymbol{T}_{image})} $)}
\State $\boldsymbol{PE} \gets \text{vf\_fun}(\boldsymbol{pos}, \boldsymbol{i})$ \Comment{Generate variant-frequency position embedding (\autoref{alg:vf_sin}): $\boldsymbol{PE} \in \mathbb{R}^{len(\boldsymbol{T}_{text} + \boldsymbol{T}_{image}) \times d_{model}}$}
\State $\boldsymbol{IE} \gets [\boldsymbol{T}_{text},  \boldsymbol{T}_{image}] \oplus \boldsymbol{PE} / d_{model}$ \Comment{Update input embeddings}
\State $\boldsymbol{T}_{next} \gets f_{VLM}(\boldsymbol{IE}, \boldsymbol{ids})$ \Comment{Predict next token}
\end{algorithmic}
\end{algorithm}

\begin{algorithm}[H]
\caption{Extract Physical Information for Image Tokens}
\label{alg:extract}
\begin{algorithmic}[1]
\Require{Input $\boldsymbol{x}, \boldsymbol{T}_{image}$}
\Ensure{Time ($\boldsymbol{t}_{day}, \boldsymbol{t}_{hour}$) and location ($\boldsymbol{lat}, \boldsymbol{lng}$) for each image tokens}
\State $\boldsymbol{t} \gets \boldsymbol{x}$ \Comment{Extract temporal information from time series input (\autoref{eq:t})}
\State $\boldsymbol{t}_{day}, \boldsymbol{t}_{hour} \gets \boldsymbol{t} // 24, \boldsymbol{t} \% 24$
\State $\boldsymbol{lat}_{image}, \boldsymbol{lng}_{image} \gets \boldsymbol{x}$ \Comment{Extract spatial information for the entire image.}
\State $\boldsymbol{lat}, \boldsymbol{lng} \gets \text{get\_center}(\boldsymbol{T}_{image}, \boldsymbol{lat}_{image}, \boldsymbol{lng}_{image})$ \Comment{ Compute center coordinates for patches}
\end{algorithmic}
\end{algorithm}

\begin{algorithm}[H]
\caption{Physics-Informed Positional Indexing}
\label{alg:position_indexing}
\begin{algorithmic}[1]
\Require{$\boldsymbol{T}_{text}, \boldsymbol{T}_{image}, \boldsymbol{t}, \boldsymbol{lat}, \boldsymbol{lng}$}
\Ensure{Physics-Informed $\boldsymbol{ids}$ }
\State $\boldsymbol{ids}_{text} \gets \text{sequential\_indexing}$ \Comment{Assign sequential indices to text tokens (\autoref{eq:sequence})}
\State $\boldsymbol{ids}_{image} \gets \text{physics-informed indexing}$ \Comment{Assign $[\boldsymbol{t}, \boldsymbol{lat}, \boldsymbol{lng}]$ to image tokens (\autoref{fig:method})}
\State $\boldsymbol{ids} \gets [\boldsymbol{ids}_{text}, \boldsymbol{ids}_{image}]$ \Comment{$\boldsymbol{ids} \in \mathbb{R}^{3\times len(\boldsymbol{T}_{text} + \boldsymbol{T}_{image})} $)}
\end{algorithmic}
\end{algorithm}

\begin{algorithm}[H]
\caption{Variant-Frequency Sinusoidal Encoding}
\label{alg:vf_sin}
\begin{algorithmic}[1]
\Require{Position $\boldsymbol{pos}$}
\Ensure{Variant-frequency position embedding $\boldsymbol{PE}$ }
\State $\boldsymbol{PE}_{text} \gets \text{standard sinusoidal function (\autoref{eq:sin})}$
\State $\boldsymbol{PE}_{image} \gets \text{variant-frequency sinusoidal function  (\autoref{eq:detail_sin})}$
\State $\boldsymbol{PE} \gets [\boldsymbol{PE}_{text}, \boldsymbol{PE}_{image}]$ 

\end{algorithmic}
\end{algorithm}

\subsection{Variant-frequency sinusoidal function}
\begin{figure}[H]
  \centering
  \includegraphics[width=\linewidth]{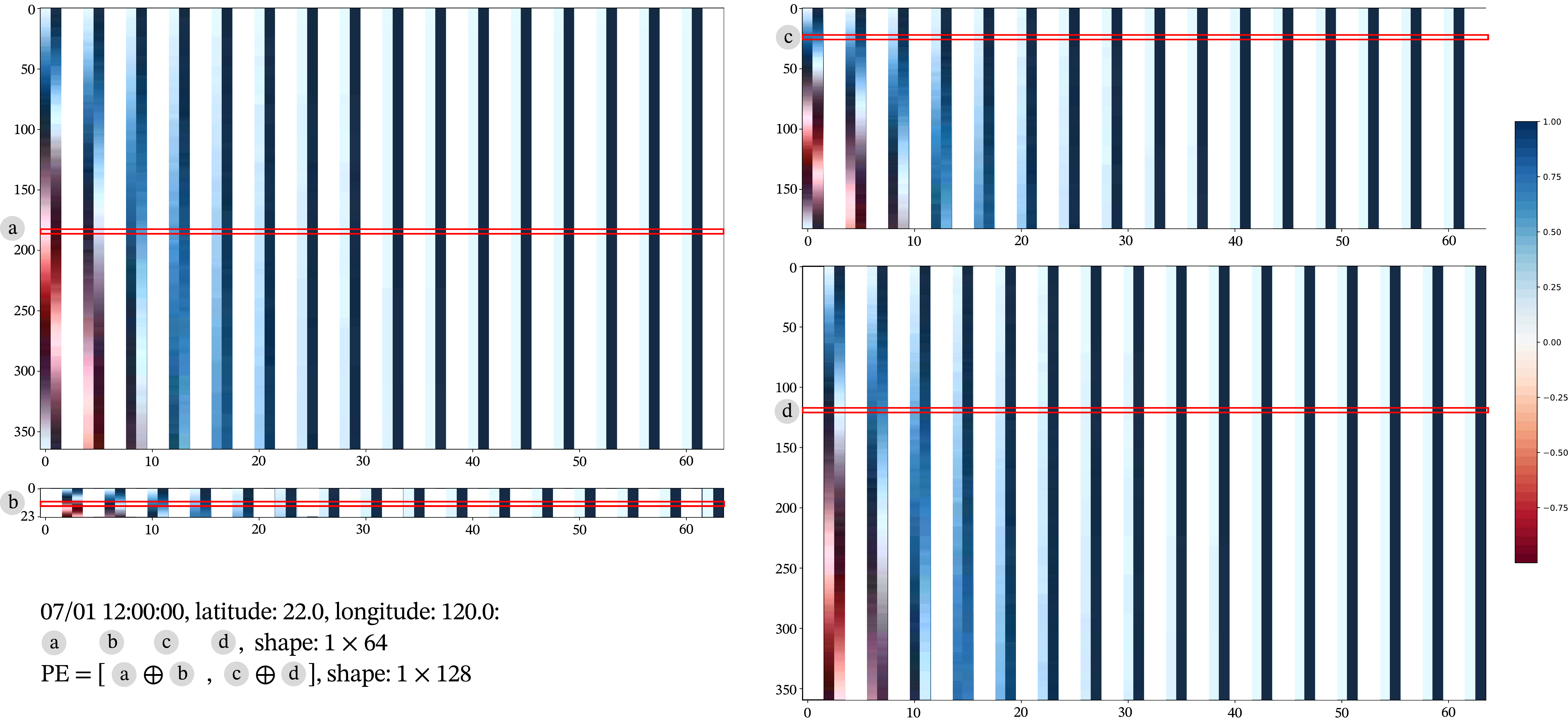}
  \caption{The 64-dimensional positional encoding for the physical variables. Each row represents the embedding vector. The final position encoding will be 128-dimensional by combining them.}
  \label{fig:pe}
\end{figure}

This section presents the complete formal definition of the variant-frequency sinusoidal function. The model dimensions are partitioned into two distinct components: temporal dimensions (first half) and spatial dimensions (latter half). Regarding the temporal dimensions, they combine the encoding of $t_{day}$ and $t_{hour}$. Similarly, for the spatial dimensions, they combine the latitude embeddings for $lat$ and the longitude embeddings for $lng$. This dimensional combination enables simultaneous representation of both temporal and spatial characteristics within the unified model framework. We also visualize the function (\autoref{fig:pe}) taking the $d_{model}=128$ as an example.

\begin{align}
\label{eq:detail_sin}
    PE_{(pos,4i)}=sin(\frac{t_{day}}{10000^{4i/d_{model}}}\times\frac{2\pi}{p_{day}}) \; if\; 4i \leq \frac{d_{model}}{2} \\
    PE_{(pos,4i+1)}=cos(\frac{t_{day}}{10000^{4i/d_{model}}}\times\frac{2\pi}{p_{day}}) \; if\; 4i+1 \leq \frac{d_{model}}{2} \nonumber \\
    PE_{(pos,4i+2)}=sin(\frac{t_{hour}}{10000^{4i/d_{model}}}\times\frac{2\pi}{p_{hour}}) \; if\; 4i+2 \leq \frac{d_{model}}{2} \nonumber \\
    PE_{(pos,4i+3)}=cos(\frac{t_{hour}}{10000^{4i/d_{model}}}\times\frac{2\pi}{p_{hour}}) \; if\; 4i+3 \leq \frac{d_{model}}{2} \nonumber \\
    PE_{(pos,4i)}=sin(\frac{lat}{10000^{4i/d_{model}-1/2}}\times\frac{2\pi}{p_{lat}}) \; if\; \frac{d_{model}}{2} < 4i \leq d_{model} \nonumber \\
    PE_{(pos,4i+1)}=cos(\frac{lat}{10000^{4i/d_{model}-1/2}}\times\frac{2\pi}{p_{lat}}) \; if\; \frac{d_{model}}{2} < 4i+1 \leq d_{model} \nonumber \\
    PE_{(pos,4i+2)}=sin(\frac{lng}{10000^{4i/d_{model}-1/2}}\times\frac{2\pi}{p_{lng}}) \; if\; \frac{d_{model}}{2} < 4i+2 \leq d_{model} \nonumber \\
    PE_{(pos,4i+3)}=cos(\frac{lng}{10000^{4i/d_{model}-1/2}}\times\frac{2\pi}{p_{hour}}) \; if\; \frac{d_{model}}{2} < 4i+3 \leq d_{model} \nonumber
\end{align}
where for temporal dimensions $p_{day}=366$ and $p_{hour}=24$, while for spatial dimensions, $p_{latitude}=180$ and $p_{longitude}=360$. $t_{day}$ is the day of the year (ranging from 0 to 365) and $t_{hour}$ represents the hour of the day (ranging from 0 to 23). $lat$ is the latitude of the image token, and $lng$ is the longitude of the image token.

\section{Implementation Details}
\label{appendix:implement}
\paragraph{VLMs training} We leverage LLama-Factory~\cite{zheng2024llamafactory} for training VLMs, utilizing PyTorch~\cite{paszke2019pytorch} on NVIDIA H800 GPUs. The optimization process employs the AdamW optimizer~\cite{loshchilov2017decoupled} with an initial learning rate of $10^{-5}$ (using the cosine scheduler), a batch size of 1, and CrossEntropy loss over 1 training epoch. 
We provide the code for the reproduction.

\paragraph{Non-vision models}
One Fits All~\cite{zhou2023one} and AutoTimes~\cite{liu2024autotimes} are implemented using their official repositories, adapted to accommodate our typhoon sequence dataset via modifications to the data loader. Configuration follows original specifications: model dimensions of 768 (One Fits All) and 512 (AutoTimes), with a batch size of 128, learning rate of $10^{-4}$, and 10 training epochs.
For other AI models, we use the publicly available platform Time-Series-Library~\cite{wu2023timesnet} to implement them. The parameters for model dimensions and number of heads are based on their implementation (512) with a batch size of 128 and a learning rate of $10^{-4}$, training epochs of 30, and patience of 10.

For domain models, the results are provided by the original paper.

\paragraph{Training Cost}
\label{appendix:cost}
We use 4×NVIDIA H800 GPUs to train the models for one epoch. 
The training time varies significantly across model sizes:
\begin{itemize}
    \item PIPE-3B: 2.1 hours
    \item PIPE-7B: 3.7 hours
    \item PIPE-32B (LoRA~\cite{hu2022lora} rank as 8): 0.7 hours
\end{itemize}
The PIPE-32B variant achieves substantial time efficiency through LoRA, which reduces trainable parameters while maintaining competitive performance (as shown in Tables~\ref{tab:scale_6} and~\ref{tab:scale_12}). This demonstrates an effective balance between model capacity and computational overhead.
For baseline AI models (including LLM-based variants like AutoTimes (OPT model)), training completes in 1 hour with a single NVIDIA RTX 4090 GPU.

\section{Dataset}
We provide a comprehensive summary of the Digital Typhoon dataset~\cite{kitamoto2023digital}.

\begin{table*}[h]
\caption{The detailed information of the Digital Typhoon dataset.}
\label{tab:typhoon}
\centering
\resizebox{\textwidth}{!}{
\begin{tabular}{p{4cm}p{7cm}} 
\toprule
& Digital Typhoon dataset \\
\midrule
Temporal coverage & 1978-2023 (present)  \\
Temporal resolution & one hour  \\
Target satellites & Himawari \\
Spatial coverage & Western North Pacific basin  \\
Spatial resolution & 5km  \\
Image coverage & 512×512 pixels (1250km from the center) \\
Spectral coverage & infrared (others on the Website)  \\
Map projection & Azimuthal equal-area projection  \\
Calibration & Recalibration \\
Data format & HDF5  \\
Best track & Japan Meteorological Agency  \\
Dataset browsing & Digital Typhoon website \\

\midrule
\bottomrule
\end{tabular}
}
\end{table*}

\section{Supplementary Results}
\subsection{Forecasting Results of More Leading Times}
\begin{figure}[h]
  \centering
  \includegraphics[width=\linewidth]{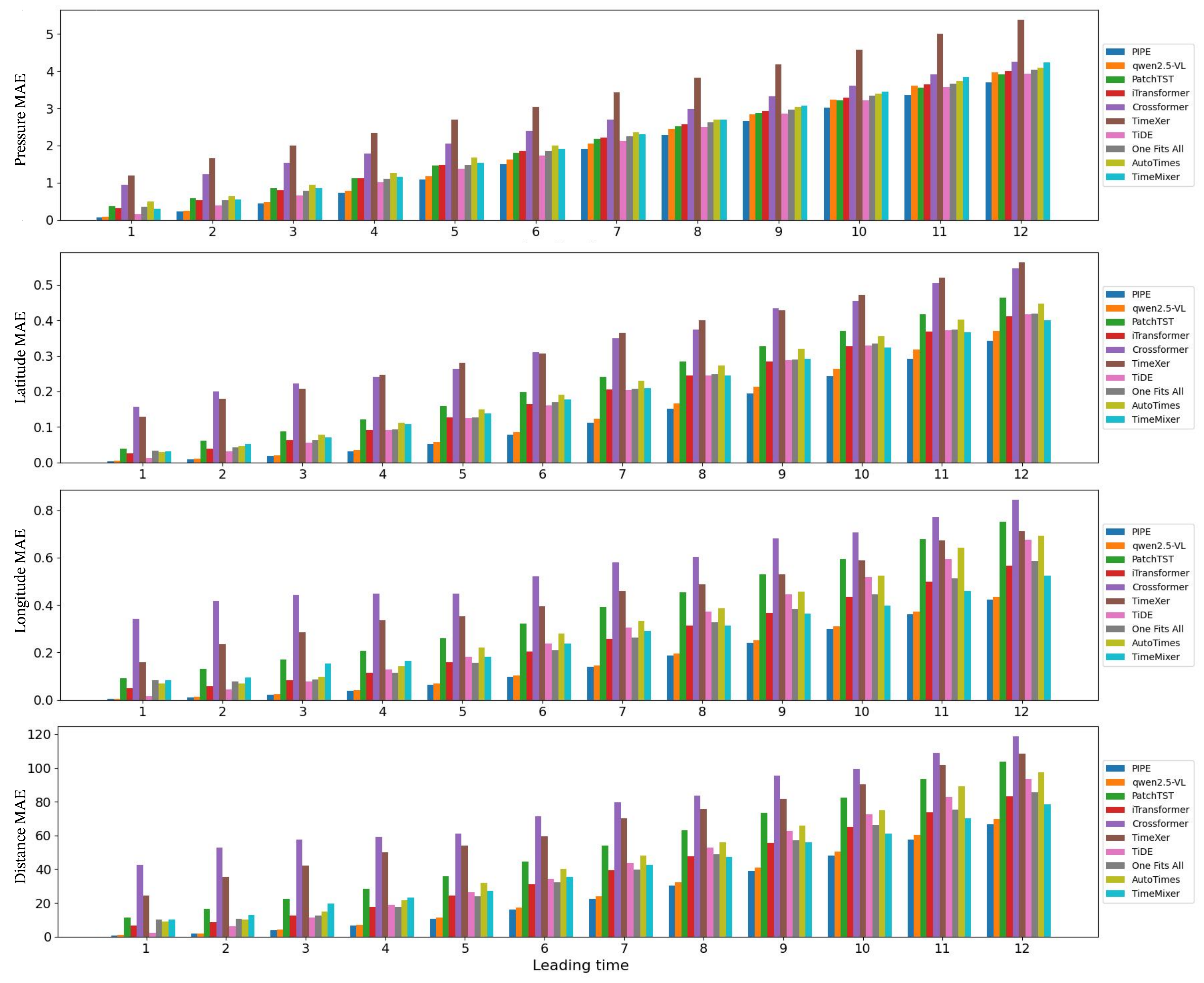}
  \caption{The performance across leading times ranging from 1 to 12 hours.}
  \label{fig:res_bar}
\end{figure}

We present additional forecasting analyses in this section. First, we list the 12-hour lead-time forecasting performance of all baseline models (\autoref{tab:12h}). Our model demonstrates state-of-the-art results across the majority of the evaluation metrics. 
Second, we list the result of the ablation study when the leading time is 12 hours. 
The consistency between results at different leading times confirms the robustness of our architectural design, demonstrating that all modules contribute meaningfully to forecasting accuracy.
Finally, we visualize the MAE for the forecasting of pressure, latitude, longitude, and distance across lead times ranging from 1 to 12 hours (\autoref{fig:res_bar}). The results confirm that our model consistently achieves the lowest MAE values at all forecast leading times. This systematic advantage over baseline models highlights the effectiveness of our model in maintaining forecasting precision as the leading time increases.

\begin{table*}[t]
\caption{Multimodal time series forecasting results (leading time is 12h).}
\label{tab:12h}
\centering
\small
\def\hlinewd#1{%
  \noalign{\ifnum0=`}\fi\hrule \@height #1 \futurelet
   \reserved@a\@xhline}
\resizebox{\textwidth}{!}{
\begin{tabular}{p{0.3cm}p{3.4cm}|cc|cc|cc|c} 
\toprule
& \multirow{2}{*}{\textbf{Models}}& \multicolumn{2}{c|}{Intensity (hPa)} &\multicolumn{2}{c|}{Latitude ($^\circ$)} &\multicolumn{2}{c|}{Longitude ($^\circ$)} & Distance (km)\\ 
& &MAE &RMSE &MAE &RMSE &MAE &RMSE & MAE\\

\midrule
\parbox[t]{3mm}{\multirow{4}{*}{\rotatebox[origin=c]{90}{domain}}} & ECMWF-HRES~\cite{documentation2020part}  & \multicolumn{2}{c|}{\multirow{3}{*}{$\setminus$}} & \multicolumn{2}{c|}{\multirow{4}{*}{$\setminus$}} & \multicolumn{2}{c|}{\multirow{4}{*}{$\setminus$}} & 44.972  \\
& PanGu~\cite{bi2023accurate}  & \multicolumn{2}{c|}{} & \multicolumn{2}{c|}{} & \multicolumn{2}{c|}{} & \underline{44.630}  \\
& GenCast~\cite{price2025probabilistic}  & \multicolumn{2}{c|}{} & \multicolumn{2}{c|}{} & \multicolumn{2}{c|}{} & \textbf{37.930}  \\ 
& TIFS~\cite{ono2019operational} & $\setminus$ & 9.061 & \multicolumn{2}{c|}{} & \multicolumn{2}{c|}{} & $\setminus$  \\ 
\midrule
\parbox[t]{3mm}{\multirow{9}{*}{\rotatebox[origin=c]{90}{w/o vision}}} & PatchTST~\cite{nie2022time} & \underline{3.917} & \textbf{5.989} & 0.465 & 0.615 & 0.751 & 0.931 & 103.818 \\
 & iTransformer~\cite{liu2023itransformer} & 4.004 & \underline{6.157} & 0.412 & 0.558 & 0.565 & 0.736 & 83.174 \\
 & Crossformer~\cite{wang2023crossformer} & 4.257 & 6.303 & 0.546 & 0.726 & 0.844 & 1.109 & 118.748 \\
 & TimeXer~\cite{wang2024timexer}) & 5.380 & 7.911 & 0.563 & 0.755 & 0.713 & 0.962 & 108.665 \\
 & TiDE~\cite{das2023long} & 3.926 & 6.080 & 0.416 & 0.561 & 0.677 & 0.850 & 93.570 \\
 & One Fits All~\cite{zhou2023one}) & 4.039 & 6.212 & 0.420 & 0.568 & 0.586 & 0.759 & 85.555 \\ 
 & AutoTimes~\cite{liu2024autotimes} & 4.086 & 6.220 & 0.448 & 0.600 & 0.692 & 0.872 & 97.244 \\ 
 & TimesNet~\cite{wu2022timesnet} & 4.798 & 7.220 & 0.892 & 1.133 & 1.796 & 2.147 & 230.376 \\
 & TimeMixer~\cite{wang2024timemixer} & 4.227 & 6.290 & 0.400 & \underline{0.533} & 0.524 & 0.685 & 78.569 \\
\midrule
\parbox[t]{3mm}{\multirow{2}{*}{\rotatebox[origin=r]{90}{vision}}} & Original paper~\cite{kitamoto2023digital} & $\setminus$ & 12.100 & \multicolumn{2}{c|}{$\setminus$} & \multicolumn{2}{c|}{$\setminus$} & $\setminus$ \\
 & Qwen-2.5-VL-3B~\cite{Qwen2.5-VL} & 3.963 & 6.599 & \underline{0.371} & 0.535 & \underline{0.435} & \underline{0.610} & 69.959 \\ 
\cmidrule(l{2em}r{0em}){1-9}
 & \textbf{PIPE-3B} & \textbf{3.855} & 6.333 & \textbf{0.359} & \textbf{0.526} & \textbf{0.411} & \textbf{0.587} & 67.114 \\
\bottomrule
\end{tabular}

}
\end{table*}

\begin{table*}[h]
\caption{The results of the ablation study (leading time is 12h).}
\label{tab:ablation_12h}
\centering
\resizebox{\textwidth}{!}{
\begin{tabular}{l|cc|cc|cc|c} 
\toprule
\multirow{2}{*}{\textbf{Models}}& \multicolumn{2}{c|}{Intensity (hPa)} &\multicolumn{2}{c|}{Latitude ($^\circ$)} &\multicolumn{2}{c|}{Longitude ($^\circ$)} & Distance (km)\\ 
&MAE &RMSE &MAE &RMSE &MAE &RMSE & MAE\\

\midrule
w/o vision & 4.120 & 6.820 & 0.372 & 0.532 & 0.436 & 0.616 & 70.138 \\
\midrule
w/o 3D indexing (using sequence) & 3.936 & 6.809 & 0.366 & 0.535 & 0.434 & 0.611 & 69.382 \\
w/o physics-informed indexing (using 3D) & 3.963 & 6.599 & 0.371 & 0.535 & 0.435 & 0.610 & 69.959 \\ 
w/o negative indexing & 4.282 & 7.017 & 0.550 & 0.806 & 0.869 & 1.306 & 124.329 \\
\midrule
w/o entire sinusoidal function & \textbf{3.827} & 6.387 & 0.364 & 0.527 & 0.416 & 0.590 & 67.904 \\
w/o variant-frequency sinusoidal function & 4.071 & 6.689 & 0.370 & 0.537 & 0.429 & 0.604 & 69.389 \\

\midrule
\textbf{PIPE-3B} & 3.855 & \textbf{6.333} & \textbf{0.359} & \textbf{0.526} & \textbf{0.411} & \textbf{0.587} & \textbf{67.114} \\
\bottomrule
\end{tabular}
}
\end{table*}

\subsection{Experiment Statistical Report}
\label{appendix:statistic}
The stability of PIPE’s forecasting performance is validated through standard deviation analysis across three random seeds, reported in Tables \autoref{tab:err_bar_6} and \autoref{tab:err_bar_12}.

\begin{table*}[h]
\caption{The mean and the standard deviation of PIPE-3B from three random seeds (leading time is 6 hours).}
\label{tab:err_bar_6}
\centering
\resizebox{\textwidth}{!}{
\begin{tabular}{l|cc|cc|cc|c} 
\toprule
\multirow{2}{*}{\textbf{Models}}& \multicolumn{2}{c|}{Intensity (hPa)} &\multicolumn{2}{c|}{Latitude ($^\circ$)} &\multicolumn{2}{c|}{Longitude ($^\circ$)} & Distance (km)\\ 
&MAE &RMSE &MAE &RMSE &MAE &RMSE & MAE\\

\midrule
PIPE-3B-seed1 & 1.503 & 2.940 & 0.085 & 0.161 & 0.095 & 0.178 & 16.364 \\
PIPE-3B-seed2 & 1.513 & 2.946 & 0.082 & 0.154 & 0.094 & 0.173 & 16.000 \\
PIPE-3B-seed3 & 1.529 & 3.059 & 0.084 & 0.161 & 0.097 & 0.181 & 16.463 \\

\midrule
PIPE-3B & 1.515 $\pm$ 0.011 & 2.981 $\pm$ 0.055 & 0.084 $\pm$ 0.001 & 0.159 $\pm$ 0.003 & 0.095 $\pm$ 0.001 & 0.178 $\pm$ 0.003 & 16.275 $\pm$ 0.200\\
\bottomrule
\end{tabular}
}
\end{table*}

\begin{table*}[h]
\caption{The mean and the standard deviation of PIPE-3B from three random seeds (leading time is 12 hours).}
\label{tab:err_bar_12}
\centering
\resizebox{\textwidth}{!}{
\begin{tabular}{l|cc|cc|cc|c} 
\toprule
\multirow{2}{*}{\textbf{Models}}& \multicolumn{2}{c|}{Intensity (hPa)} &\multicolumn{2}{c|}{Latitude ($^\circ$)} &\multicolumn{2}{c|}{Longitude ($^\circ$)} & Distance (km)\\ 
&MAE &RMSE &MAE &RMSE &MAE &RMSE & MAE\\

\midrule
PIPE-3B-seed1 & 3.840 & 6.295 & 0.362 & 0.530 & 0.412 & 0.590 & 67.432 \\
PIPE-3B-seed2 & 3.831 & 6.281 & 0.355 & 0.523 & 0.405 & 0.578 & 66.402 \\
PIPE-3B-seed3 & 3.893 & 6.425 & 0.360 & 0.527 & 0.415 & 0.592 & 67.506 \\

\midrule
PIPE-3B & 3.855 $\pm$ 0.027 & 6.333 $\pm$ 0.065 & 0.359 $\pm$ 0.003 & 0.526 $\pm$ 0.003 & 0.411 $\pm$ 0.004 & 0.587 $\pm$ 0.006 & 67.114 $\pm$ 0.050\\
\bottomrule
\end{tabular}
}
\end{table*}

\subsection{Scaling Behavior}
To evaluate the impact of model size on performance, we conduct experiments across three variants: PIPE-3B, PIPE-7B, and PIPE-32B (with LoRA rank as 8). As demonstrated in Tables~\ref{tab:scale_6} and~\ref{tab:scale_12}, the largest model, PIPE-32B, yields performance improvements, even when leveraging LoRA.

\begin{table*}[h]
\caption{The results of PIPE-3B, PIPE-7B, and PIPE-32B (with LoRA) with the lead time of 6 hours.}
\label{tab:scale_6}
\centering
\resizebox{\textwidth}{!}{
\begin{tabular}{l|cc|cc|cc|c} 
\toprule
\multirow{2}{*}{\textbf{Models}}& \multicolumn{2}{c|}{Intensity (hPa)} &\multicolumn{2}{c|}{Latitude ($^\circ$)} &\multicolumn{2}{c|}{Longitude ($^\circ$)} & Distance (km)\\ 
&MAE &RMSE &MAE &RMSE &MAE &RMSE & MAE\\

\midrule
PIPE-3B & 1.515 & 2.981 & 0.084 & 0.159 & 0.095 & 0.178 & 16.275 \\
PIPE-7B & 1.505 & 2.918 & 0.088 & 0.166 & 0.102 & 0.184 & 17.194 \\
PIPE-32B & 1.505 & 2.874 & 0.079 & 0.153 & 0.097 & 0.182 & 15.980 \\

\bottomrule
\end{tabular}
}
\end{table*}

\begin{table*}[t]
\caption{The results of PIPE-3B, PIPE-7B, and PIPE-32B (with LoRA) with the leading time of 12 hours.}
\label{tab:scale_12}
\centering
\resizebox{\textwidth}{!}{
\begin{tabular}{l|cc|cc|cc|c} 
\toprule
\multirow{2}{*}{\textbf{Models}}& \multicolumn{2}{c|}{Intensity (hPa)} &\multicolumn{2}{c|}{Latitude ($^\circ$)} &\multicolumn{2}{c|}{Longitude ($^\circ$)} & Distance (km)\\ 
&MAE &RMSE &MAE &RMSE &MAE &RMSE & MAE\\

\midrule
PIPE-3B & 3.855 & 6.333 & 0.359 & 0.526 & 0.411 & 0.587 & 67.114 \\
PIPE-7B & 3.861 & 6.325 & 0.371 & 0.540 & 0.435 & 0.609 & 69.933 \\
PIPE-32B & 3.695 & 6.029 & 0.342 & 0.510 & 0.423 & 0.610 & 66.725 \\

\bottomrule
\end{tabular}
}
\end{table*}

\subsection{Showcase}
\begin{figure}[h]
  \centering
  \includegraphics[width=\linewidth]{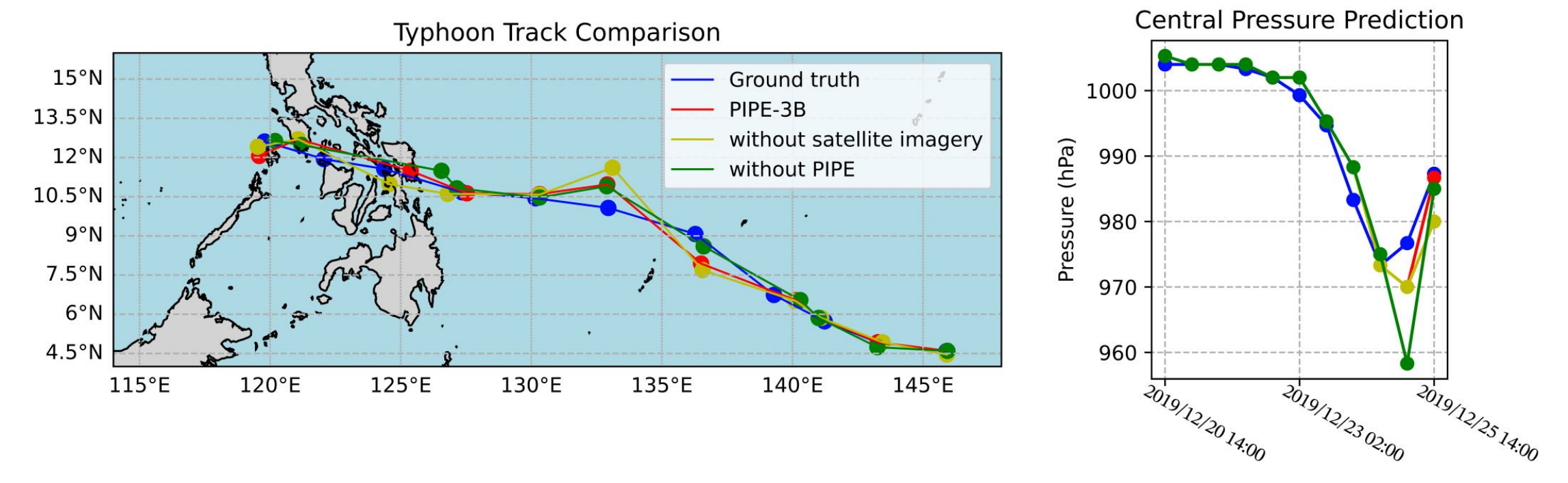}
  \caption{The results of Typhoon Phanfone comparison between PIPE, removing satellite images, and removing PIPE. The leading time is 12 hours and the time gap between neighbouring dots is 12 hours.}
  \label{fig:showcase}
\end{figure}

We present a prediction showcase (\autoref{fig:showcase}) to compare our method with the methods that remove satellite imagery and PIPE on the Typhoon Phanfone. PIPE achieves more accurate track forecasting and intensity forecasting.

\subsection{Attention Analysis}
\begin{figure}[H]
  \centering
  \includegraphics[width=\linewidth]{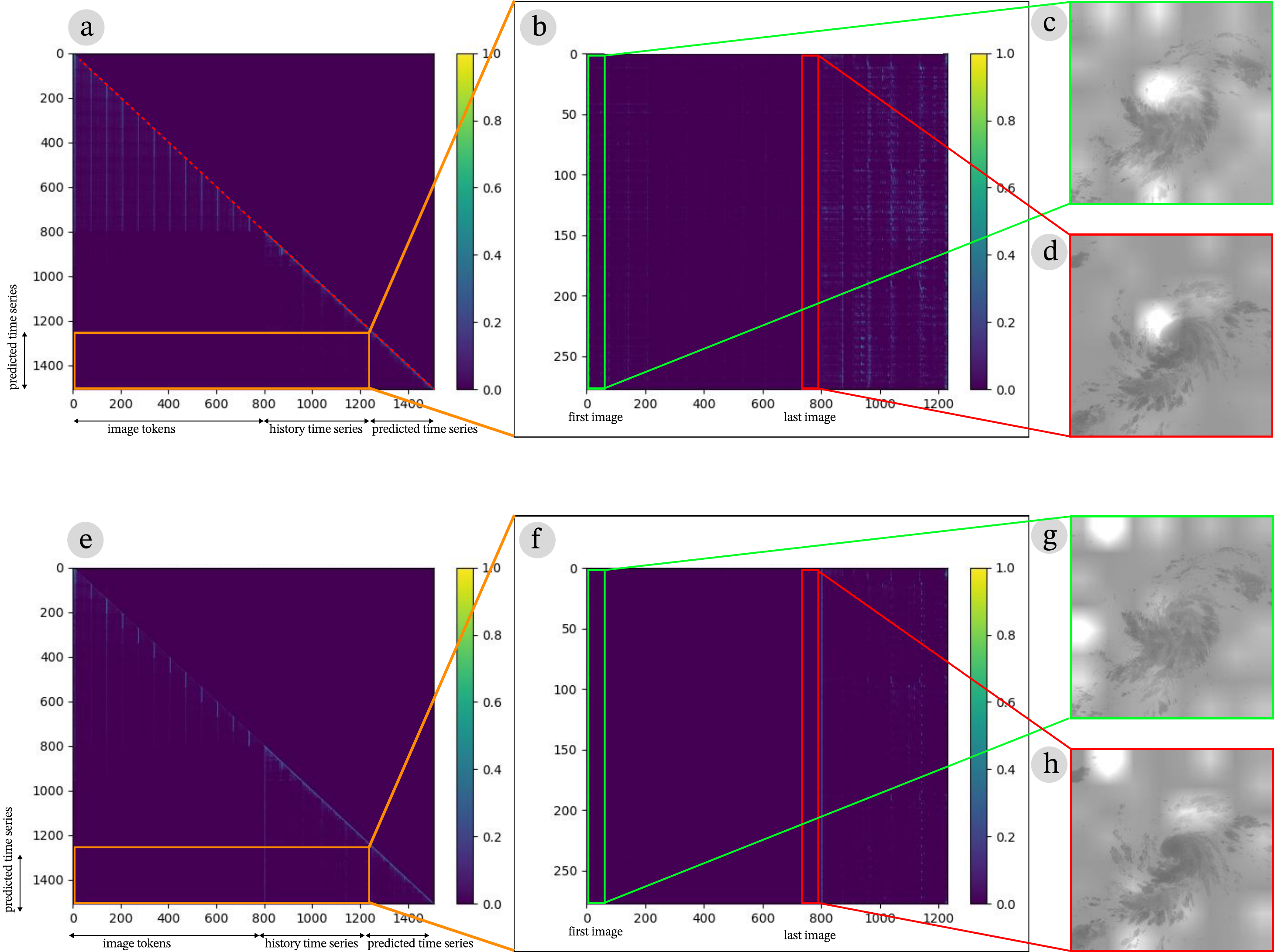}
  \caption{Visualization of attention (normalized to 0-1 in each step) from the penultimate layer of the PIPE model (top) and the Qwen2.5-VL-3B model (bottom), averaged across attention heads. (a) \& (e) The entire attention matrix. (b) \& (f) The attention matrix of predicted tokens' attention on the input tokens, including image tokens and history time series tokens. (c) \& (g) Attention of predicted tokens on the first input image. (d) \& (h) Attention of predicted tokens on the last input image.}
  \label{fig:attention_layer}
\end{figure}

We compare the attention from the penultimate layer (\autoref{fig:attention_layer}) with averaging across the head dimension of PIPE-3B and Qwen2.5-VL-3B. It reveals distinct attention patterns. Qwen2.5-VL-3B exhibits an obvious bias toward the initial tokens of the historical time series, as evidenced by an obvious vertical line at 800th input tokens ((e) \& (f)). In contrast, our PIPE model allocates greater attention to both the image tokens and the historical time series tokens. Notably, PIPE’s attention on image patches is concentrated on the typhoon region (e.g., central cloud structure), whereas Qwen2.5-VL-3B’s attention appears diffuse and unstructured across the image. These differences in attention mechanisms likely contribute to PIPE’s better forecasting accuracy.

\begin{figure}[H]
  \centering
  \includegraphics[width=\linewidth]{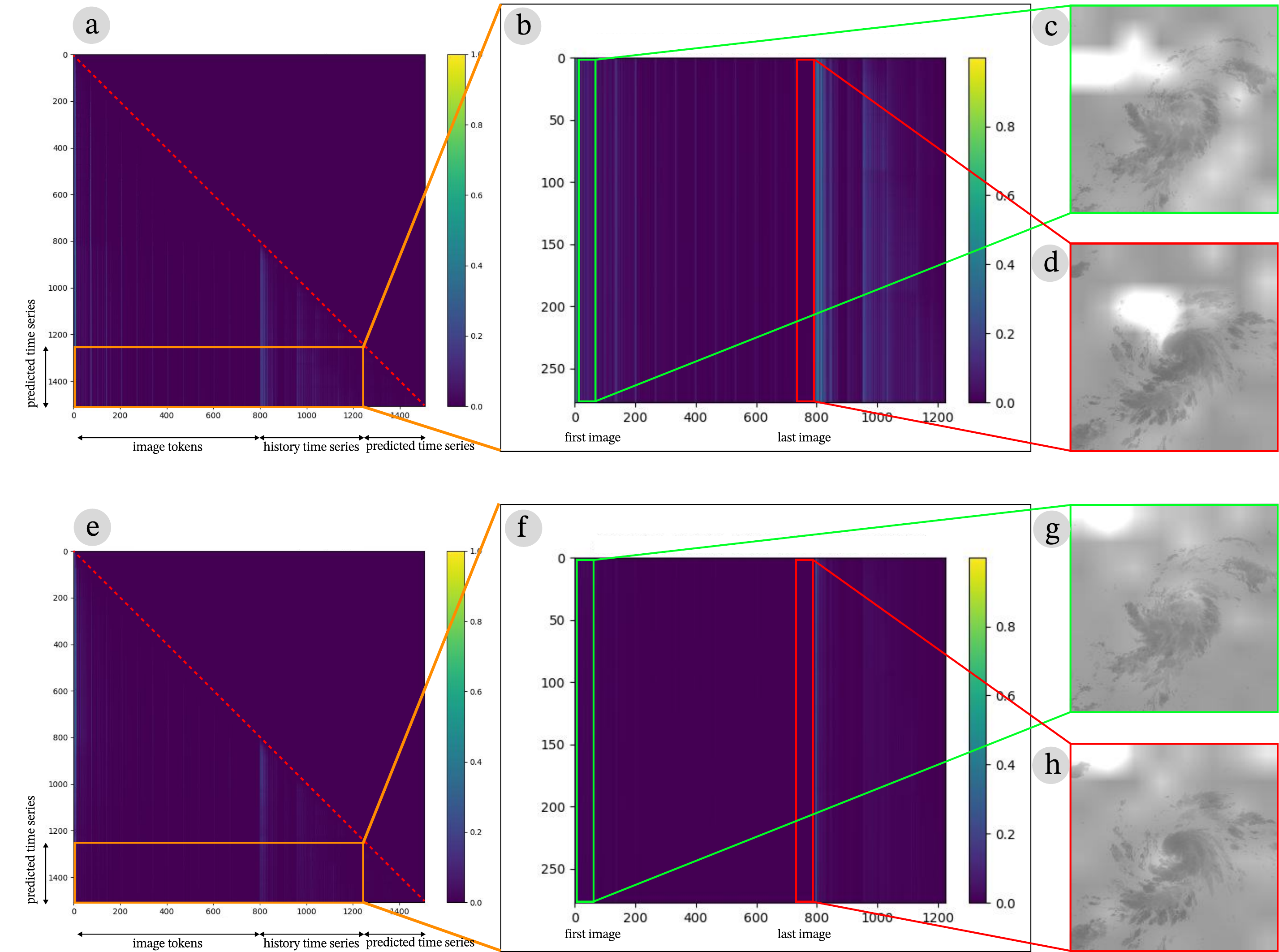}
  \caption{Visualization of attention (normalized to 0-1 in each step) using Attention Rollout of the PIPE model (top) and the Qwen2.5-VL-3B model (bottom), averaged across attention heads. (a) \& (e) The entire attention matrix. (b) \& (f) The attention matrix of predicted tokens' attention on the input tokens, including image tokens and history time series tokens. (c) \& (g) Attention of predicted tokens on the first input image. (d) \& (h) Attention of predicted tokens on the last input image.}
  \label{fig:attention_rollout}
\end{figure}

We also compare the attention using Attention Rollout~\cite{abnar2020quantifying} (\autoref{fig:attention_rollout}) with averaging across the head dimension of PIPE-3B and Qwen2.5-VL-3B. 
It also demonstrates that our model allocates more reasonable attention to image tokens and historical time series tokens. Furthermore, our model's attention on image patches is focused specifically on the typhoon region, whereas Qwen2.5-VL-3B's attention appears biased across the image.

\section{Broader Impact}
\label{appendix:broader}
We introduce a novel multimodal time series forecasting task that integrates satellite imagery with temporal data for capturing complex spatio-temporal dependencies. This approach leverages the complementary strengths of temporal time series data and spatially rich visual inputs, enabling models to go beyond the limitations of traditional univariate, multivariate, or single-modality methods. 
To address the inherent challenges of integrating satellite imagery into time series forecasting, we propose a physics-informed positional encoding. This technique incorporates physical information derived from satellite data, such as geospatial coordinates, to enhance the model's ability to reason about spatial and temporal dependencies. This innovation is particularly relevant for applications where visual inputs carry critical physical context, including climate modeling, urban planning, and agricultural forecasting. 
The broader impact of this work lies in its ability to bridge the gap between traditional forecasting methods and real-world complexities that often include spatial and physical components. By incorporating satellite imagery and physics-informed encoding, this method has potential benefits across a wide range of scientific and practical domains.

\section{Limitation}
\label{sec:limitation}
While the integration of satellite imagery improves forecasting accuracy, it increases the computational complexity needed to process high-resolution images.
To address these limitations, future work will focus on improving the efficiency of integrating vision data into forecasting models to enable longer input sequences and extended forecasting horizons for VLMs.
Furthermore, we will explore the incorporation of physical laws or constraints. Beyond embedding physical information, integrating domain-specific physical principles or environmental constraints could improve the model's interpretability and robustness.


\newpage
\section*{NeurIPS Paper Checklist}
\begin{enumerate}

\item {\bf Claims}
    \item[] Question: Do the main claims made in the abstract and introduction accurately reflect the paper's contributions and scope?
    \item[] Answer: \answerYes{} 
    \item[] Justification: We state our claim and contributions in our abstract and introduction.

\item {\bf Limitations}
    \item[] Question: Does the paper discuss the limitations of the work performed by the authors?
    \item[] Answer: \answerYes{} 
    \item[] Justification: We discuss the limitation in \autoref{sec:limitation}.

\item {\bf Theory assumptions and proofs}
    \item[] Question: For each theoretical result, does the paper provide the full set of assumptions and a complete (and correct) proof?
    \item[] Answer: \answerNA{} 
    \item[] Justification: We do not include theoretical results.

    \item {\bf Experimental result reproducibility}
    \item[] Question: Does the paper fully disclose all the information needed to reproduce the main experimental results of the paper to the extent that it affects the main claims and/or conclusions of the paper (regardless of whether the code and data are provided or not)?
    \item[] Answer: \answerYes{} 
    \item[] Justification: We disclose the implementation details in \autoref{sec:experment} and \autoref{appendix:implement} for reproducing the results. We also provide the code and data in supplementary materials for reproduction.

\item {\bf Open access to data and code}
    \item[] Question: Does the paper provide open access to the data and code, with sufficient instructions to faithfully reproduce the main experimental results, as described in supplemental material?
    \item[] Answer: \answerYes{} 
    \item[] Justification: We provide the code and data in the supplementary materials.

\item {\bf Experimental setting/details}
    \item[] Question: Does the paper specify all the training and test details (e.g., data splits, hyperparameters, how they were chosen, type of optimizer, etc.) necessary to understand the results?
    \item[] Answer: \answerYes{} 
    \item[] Justification: Please refer to \autoref{sec:experment} and \autoref{appendix:implement} for the training and test details. We introduce settings, training hyperparameters, optimizer, etc.

\item {\bf Experiment statistical significance}
    \item[] Question: Does the paper report error bars suitably and correctly defined or other appropriate information about the statistical significance of the experiments?
    \item[] Answer: \answerYes{} 
    \item[] Justification: We report standard deviations with three random seeds in \autoref{appendix:statistic}.

\item {\bf Experiments compute resources}
    \item[] Question: For each experiment, does the paper provide sufficient information on the computer resources (type of compute workers, memory, time of execution) needed to reproduce the experiments?
    \item[] Answer: \answerYes{} 
    \item[] Justification: We provide the computation cost in \autoref{appendix:implement}.
    
\item {\bf Code of ethics}
    \item[] Question: Does the research conducted in the paper conform, in every respect, with the NeurIPS Code of Ethics \url{https://neurips.cc/public/EthicsGuidelines}?
    \item[] Answer: \answerYes{} 
    \item[] Justification: We have checked the NeurIPS Code of Ethics and our research conforms with it.

\item {\bf Broader impacts}
    \item[] Question: Does the paper discuss both potential positive societal impacts and negative societal impacts of the work performed?
    \item[] Answer: \answerYes{} 
    \item[] Justification: We discuss the broader impacts of our work in \autoref{appendix:broader}, including the impacts on other domains and scenarios.
    
\item {\bf Safeguards}
    \item[] Question: Does the paper describe safeguards that have been put in place for responsible release of data or models that have a high risk for misuse (e.g., pretrained language models, image generators, or scraped datasets)?
    \item[] Answer: \answerNA{} 
    \item[] Justification: Our paper poses no such risks.

\item {\bf Licenses for existing assets}
    \item[] Question: Are the creators or original owners of assets (e.g., code, data, models), used in the paper, properly credited and are the license and terms of use explicitly mentioned and properly respected?
    \item[] Answer: \answerYes{} 
    \item[] Justification: We properly cite the assets including dataset and code in \autoref{sec:experment} and \autoref{appendix:implement}.

\item {\bf New assets}
    \item[] Question: Are new assets introduced in the paper well documented and is the documentation provided alongside the assets?
    \item[] Answer: \answerYes{} 
    \item[] Justification: We provide the readme file for our code in the supplementary materials. 

\item {\bf Crowdsourcing and research with human subjects}
    \item[] Question: For crowdsourcing experiments and research with human subjects, does the paper include the full text of instructions given to participants and screenshots, if applicable, as well as details about compensation (if any)? 
    \item[] Answer: \answerNA{} 
    \item[] Justification: Our work does not involve crowdsourcing nor research with human subjects.

\item {\bf Institutional review board (IRB) approvals or equivalent for research with human subjects}
    \item[] Question: Does the paper describe potential risks incurred by study participants, whether such risks were disclosed to the subjects, and whether Institutional Review Board (IRB) approvals (or an equivalent approval/review based on the requirements of your country or institution) were obtained?
    \item[] Answer: \answerNA{} 
    \item[] Justification: Our paper does not involve crowdsourcing nor research with human subjects.

\item {\bf Declaration of LLM usage}
    \item[] Question: Does the paper describe the usage of LLMs if it is an important, original, or non-standard component of the core methods in this research? Note that if the LLM is used only for writing, editing, or formatting purposes and does not impact the core methodology, scientific rigorousness, or originality of the research, declaration is not required.
    \item[] Answer: \answerNA{} 
    \item[] Justification: We do not use LLMs for the core method development.

\end{enumerate}

\end{document}